\documentclass{article}

\usepackage{arxiv}

\usepackage[utf8]{inputenc} 
\usepackage[T1]{fontenc}    
\usepackage{hyperref}       
\usepackage{url}            
\usepackage{booktabs}       
\usepackage{amsfonts}       
\usepackage{nicefrac}       
\usepackage{microtype}      
\usepackage{lipsum}		
\usepackage{graphicx}
\usepackage[numbers]{natbib}
\usepackage{doi}
\usepackage{amsmath}
\usepackage{amssymb}
\usepackage{graphicx}
\usepackage{caption}
\usepackage{subfig}
\usepackage{xcolor}
\usepackage{makecell}
\usepackage{multirow}
\usepackage{multicol}
\usepackage{enumitem}
\usepackage{tikz}

\usepackage{mathtools} 

\hypersetup{
    colorlinks,
    linkcolor={blue!80!black},
    citecolor={blue!80!black},
    urlcolor={blue!80!black}
}

\usepackage{etoolbox}
\def\intdomd{\Omega_\delta}
\def\uelast{u_l}
\def\uperid{u}

\title{ML-based identification of the interface regions for coupling local and nonlocal models}

\date{} 					

\author{{Noujoud Nader$^*$} \\
	Louisiana State University\\
	Baton Rouge, 70803, LA, USA \\
	\texttt{nnader@lsu.edu} \\
	\And
	{Patrick Diehl} \\
	Louisiana State University\\
	Baton Rouge, 70803, LA, USA \\
	\texttt{pdiehl@cct.lsu.edu} \\
         \And
        {Marta D’Elia} \\
	Pasteur Labs\\
	Brooklyn, NY 11205 \\
	\texttt{marta.delia@simulation.science} \\
  \And
        {Christian Glusa} \\
	Center for Computing Research, Sandia National Laboratories\\
	Albuquerque, NM 87321\\
	\texttt{caglusa@sandia.gov} \\
   \And
        {Serge Prudhomme} \\
    Department of Mathematics and Industrial Engineering, Polytechnique Montr\'eal,\\
	C.P. 6079, succ. Centre-ville Montr\'eal, QC H3C 3A7, Canada\\
	\texttt{serge.prudhomme@polymtl.ca} \\
 }

\renewcommand{\shorttitle}{ML-based identification of the interface regions for coupling local and nonlocal models}

\begin{document}
\maketitle
\begin{abstract}
Local-nonlocal coupling approaches provide a means to combine the computational efficiency of local models and the accuracy of nonlocal models. However, the coupling process can be challenging, requiring expertise to identify the interface between local and nonlocal regions.
This study introduces a machine learning-based approach to automatically detect the regions in which the local and nonlocal models should be used in a coupling approach. 
This identification process takes as input the loading functions evaluated at the grid points and provides as output the selected model at those points. Training of the networks is based on datasets provided by classes of loading functions for which reference coupling configurations are computed using accurate coupled solutions, where accuracy is measured in terms of the relative error between the solution to the coupling approach and the solution to the nonlocal model. 
We study two approaches that differ from one another in terms of the data structure. The first approach, referred to as the full-domain input data approach, inputs the full load vector and outputs a full label vector. In this case, the classification process is carried out globally. The second approach consists of a window-based approach, where loads are preprocessed and partitioned into windows and the problem is formulated as a node-wise classification approach in which the central point of each window is treated individually.
The classification problems are solved via deep learning algorithms based on convolutional neural networks. The performance of these approaches is studied on one-dimensional numerical examples using F1-scores and accuracy metrics. In particular, it is shown that the windowing approach provides promising results, achieving an accuracy of 0.96 and an F1-score of 0.97. These results underscore the potential of the approach to automate coupling processes, leading to more accurate and computationally efficient solutions for material science applications.
\end{abstract}

\keywords{coupling methods, local and nonlocal models, loading functions, machine learning, convolutional neural networks}

\section{Introduction}
Recent advances in engineering mechanics have seen an increasing emphasis on coupling nonlocal models with classical local models, an approach driven by the need to address the excessive computational costs of high fidelity models, such as nonlocal equations, for modeling and simulation of complex behaviors in materials science. Nonlocal models, such as nonlocal diffusion and peridynamics, offer a more comprehensive representation of phenomena that are not adequately captured by classical Partial Differential Equations (PDEs). These phenomena include fracture and cracks in mechanics~\cite{diehl2022comparative}. The scope of nonlocal models extends beyond diffusion and mechanics, finding applications in diverse areas like subsurface transport~\cite{benson2000application,suzuki2023fractional,katiyar2014peridynamic, katiyar2020general,xu2022machine}, phase transitions~\cite{bates1999integrodifferential, chen2000nonlocal, dayal2006kinetics} and image processing~\cite{d2021bilevel,lou2010image}.

However, the implementation and simulation of nonlocal models present several challenges. These include the complexity of handling nonlocal boundary conditions~\cite{diehl2019review} and the high computational costs associated with their numerical solutions. To address these challenges, coupling between local and nonlocal models approach has been proposed. Coupling methods aim to merge the computational efficiency of PDEs with the accuracy of nonlocal models, particularly in cases where nonlocal effects are confined to specific domain areas, and the rest of the system can be effectively described by a PDE.

For a general overview of local-nonlocal coupling methods, classified as either force-based or energy-based formulations, we refer to the recent survey~\cite{d2021review}.  Our focus is on the coupling of the bond-based peridynamic model with the classical linear elasticity continuum model, using a force-based coupling formulation. In this context, various coupling approaches have been proposed; these are based on matching the displacements (the Matching Displacement Coupling Method) or the stresses (the Matching Stress Coupling Method) either over an overlap region or a sharp interface obtained by shrinking the size of the nonlocal horizon when transitioning from a nonlocal to a local region. The latter approach is known as the variable horizon coupling method (VHCM); the interested reader can find an extended description in~\cite{diehl2022coupling}.

A key aspect of the coupling process is that it requires domain expertise to effectively choose the interface between the local and nonlocal regions. In this work, we propose to circumvent this challenge by letting machine learning detect such an interface region using a supervised approach. Specifically, the use of machine learning involves training neural network architectures on datasets containing local-nonlocal coupling regions that yield a certain accuracy of the coupled solution. Once trained, the networks can automatically identify the regions of the computational domain where nonlocal effects are likely to be significant, solely based on external loading information. This approach represents a significant advance, as it automates the identification of regions by removing the expert-in-the-loop.

Although to the best of our knowledge the use of machine learning for local-nonlocal coupling has never been pursued before, machine learning for nonlocal simulations and/or computational mechanics is not a new concept. In what follows, we briefly review the most relevant research. We first focus on machine learning for peridynamics. Haghighat et al.~\cite{haghighat2021nonlocal} presented a nonlocal physics-informed framework with the peridyanmic differential operator~\cite{madenci2016peridynamic} using deep learning. Kim et al.~\cite{kim2019peri} presented \textit{Peri-net} to analyze crack patterns using Deep Neural Networks. Nguyen et al.~\cite{nguyen2023peridynamic} presented peridynamic-based machine learning approach for up to two-dimensional structures. Xu et al.~\cite{XU2021114062}  presented a machine-learning-based framework for peridynamic material models with physical constraints. Concerning nonlocal models, the following literature is available. Tao et al.~\cite{tao2018nonlocal} investigated nonlocal neural networks, nonlocal diffusion, and nonlocal modeling. Feng et al.~\cite{feng2020stochastic} analyzed stochastic nonlocal damage using machine learning. Zhou et al.~\cite{zhou2021learning} learned nonlocal constitutive models using neural networks. You et al.~\cite{you2021data,You2021,you2022data} used data-driven learning of nonlocal physics from high-fidelity synthetic data. In~\cite{de2023machine}, de Moraes et al.\ used machine learning to predict the evolution of nonlocal micro-structural defects in crystalline materials.
Lal et al.~\cite{lal2023prediction} predicted nonlocal elasticity parameters using machine learning and molecular dynamics simulations. 

Concerning local models, we restrict ourselves to classical continuum mechanics. For a review of machine learning and data-miming approaches for continuum material mechanics, we refer the reader to~\cite{bock2019review}; whereas for computational solid mechanics, we refer to~\cite{kumar2022machine}. A machine-learning approach for physics-based material models for multiscale solid mechanics was presented in~\cite{rocha2023machine}. 

Finally, the use of machine learning for coupling methods (not involving nonlocal or peridynamics models) is also not new; in fact, Raymond~\cite{raymond2020combining} used machine learning to model coupled solid and fluid mechanics with mesh-free methods.

The paper is organized as follows. Section~\ref{sec:models} briefly introduces the models, coupling approaches, and discretization. In Section~\ref{sec:methodology} the methodology and the ML model which is the convolution neural network (CNN) are described. Section~\ref{sec:data_generation} covers the data generation. In Section~\ref{sec:numerical:results}, we present numerical results for two cases using multiple node classification in Subsection~\ref{sec:numerical:results:fulldomain} and the main case of this work using node-wise classification in Subsection~\ref{sec:numerical:results:classification}. Finally, Section~\ref{sec:conclusion} concludes the paper.

\section{Description of the models and coupling approach}
\label{sec:models}
The model problem that we consider in this work deals with the quasi-static simulation of a one-dimensional bar with mixed boundary conditions at the two extremities. The deformations within the bar are assumed to be small enough to be described by linear elasticity. This section briefly describes the local and nonlocal models: linear elasticity and  linearized bond-based peridynamics~\cite{silling2000reformulation}. We present the continuous formulations first and then provide their discretizations using finite differences for the local model and a collocation approach for peridynamics.

\subsection{Classical linear elasticity model}

We assume a one-dimensional bar of length $\ell$ occupying the open domain $\Omega=(0,\ell) \subset \mathbb{R}$. Denoting the closure of $\Omega$ as $\bar{\Omega}=[0,\ell]$, we seek the displacement $\uelast = \uelast(x)$, $\forall x \in \bar{\Omega}$, such that:
\begin{align}
\label{eq:1dlinearelasticity}
- EA \uelast''(x) = f_b(x), &\quad \forall x \in \Omega, \\
\label{eq:Dirichlet}
\uelast(x) = 0, &\quad \text{at}\ x=0,\\
\label{eq:Neumann}
EA\uelast'(x) = g, &\quad \text{at}\ x=\ell,
\end{align}
where $E$ is the modulus of elasticity, $A$ denotes the cross-sectional area of the bar, $f_b=f_b(x)$ is the external body force density (per unit length), and $g\in \mathbb{R}$ is a traction force applied at $x=\ell$. We assume that $f_b$ is chosen smoothly enough so that the solution $\uelast$ can be differentiated as many times as needed. For the sake of simplicity in the presentation, but without loss of generality, we assume a constant Young's modulus~$E$ and a constant cross-sectional area~$A$ for the bar. The model is presented here with a homogeneous Dirichlet boundary condition at $x=0$ and a Neumann condition at $x=\ell$, but we shall also consider examples with homogeneous Dirichlet conditions at both ends in the numerical experiments.

\subsection{Linearized bond-based peridynamic model}

The static peridynamic equation~\cite{silling2000reformulation} reads in one dimension as
\begin{equation}
\label{eq:linearizedperidynamics}
- \int_{H_\delta(x)} \kappa \frac{\uperid(y) - \uperid(x)}{|y-x|} dy = f_b(x),
\end{equation}
where the so-called horizon $\delta>0$ determines the neighborhood of point $x$
\begin{align*}
    H_\delta(x) = \{ y \in  \mathbb{R};\ \vert y - x \vert < \delta\}\text{.}
\end{align*}
In addition, $\kappa=\kappa(x)$ is the peridynamic stiffness parameter evaluated at point~$x$ and $u$ is the displacement for the deformed configuration. Following~\cite{diehl2022coupling}, the peridynamic stiffness parameter $\kappa$ is chosen such that the solution to the linearized peridynamic bond-based model is compatible with that to the continuum local model, i.e.,
\begin{equation}
\frac{\kappa\delta^2}{2} = EA, 
\end{equation}
or, equivalently,
\begin{equation}
\label{eq:kappa:variable}
\kappa = \frac{2EA}{\delta^2}.
\end{equation}
As far as the boundary conditions are concerned, we consider the same local boundary conditions~\eqref{eq:Dirichlet} and~\eqref{eq:Neumann}, which are imposed using the variable horizon method described below. The solution to this problem will serve as the reference solution. For more details about boundary conditions for the peridynamic model, we refer the reader to, e.g., \cite{DEliaNeumann2020,prudhomme2020treatment}.

\subsection{Variable horizon coupling approach}

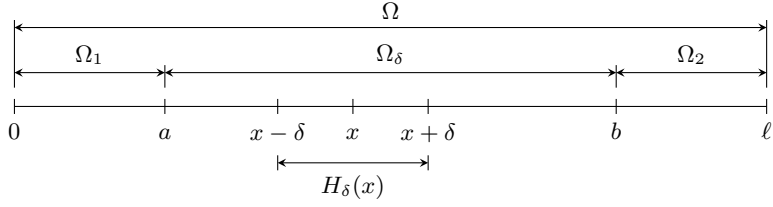
\begin{figure}[tb]
\centering
\small
\begin{tikzpicture}
\draw (0,0) -- (10.0,0);
\draw (0,-0.1) -- (0,0.1);
\draw (2,-0.1) -- (2,0.1);
\draw (3.5,-0.1) -- (3.5,0.1);
\draw (4.5,-0.1) -- (4.5,0.1);
\draw (5.5,-0.1) -- (5.5,0.1);
\draw (8.0,-0.1) -- (8.0,0.1);
\draw (10.0,-0.1) -- (10.0,0.1);
\node[above] at (0.0,-0.55) {$0$};
\node[above] at (2.0,-0.55) {$a$};
\node[above] at (3.5,-0.59) {$x-\delta$};
\node[above] at (4.5,-0.55) {$x$};
\node[above] at (5.5,-0.59) {$x+\delta$};
\node[above] at (8.0,-0.55) {$b$};
\node[above] at (10.0,-0.55) {$\ell$};
\draw[arrows=<->, >=stealth]  (0.0,1.05) -- (10.0,1.05);
\draw  (0.0,0.25) -- (0.0,1.15);
\draw  (10.0,0.25) -- (10.0,1.15);
\node[above] at (5.00,1.05) {$\Omega$};
\draw[arrows=<->, >=stealth]  (2.0,0.45) -- (8.0,0.45);
\draw  (2.0,0.25) -- (2.0,0.55);
\draw  (8.0,0.25) -- (8.0,0.55);
\node[above] at (5.00,0.45) {$\intdomd$};
\draw[arrows=<->, >=stealth]  (0.0,0.45) -- (2.0,0.45);
\node[above] at (1.00,0.45) {$\Omega_1$};
\draw[arrows=<->, >=stealth]  (8.0,0.45) -- (10.0,0.45);
\node[above] at (9.00,0.45) {$\Omega_2$};
\draw[arrows=<->, >=stealth]  (3.5,-0.75) -- (5.5,-0.75);
\draw  (3.5,-0.65) -- (3.5,-0.85);
\draw  (5.5,-0.65) -- (5.5,-0.85);
\node[below] at (4.5,-0.8) {$H_\delta(x)$};
\end{tikzpicture}
\caption{Definition of the computational domains for the coupled model. Adapted from author's previous work~\cite{diehl2022coupling}.}
\label{Fig:coupleddomains}
\end{figure}

We consider in this work the Variable Horizon Coupling Method (VHCM) as introduced in~\cite{diehl2022coupling}. A similar coupling approach was proposed in~\cite{silling2015variable, NIKPAYAM2019308}. For a broader review of coupling methods, including optimization-based and energy-based coupling methods, we refer to~\cite{d2019review}. As shown in Figure~\ref{Fig:coupleddomains}, we define the region in which we shall use the nonlocal model~\eqref{eq:linearizedperidynamics} as $\Omega_\delta = (a,b)$ and the region for the local model~\eqref{eq:1dlinearelasticity} as $\Omega_e = \Omega \backslash \bar{\Omega}_\delta$. We suppose here that $\Omega_e$ is decomposed into the two subdomains $\Omega_1 = (0,a)$ and $\Omega_2=(b,\ell)$ so that the end points of the bar are both adjacent to a domain governed by the local model. The main idea of the method is to make the horizon~$\delta$ decrease to zero as one approaches the interfaces~$x=a$ and~$x=b$ so as to avoid the need to introduce an overlapping region around the interfaces. In this section, we describe the main ingredients of the VHCM method.

\begin{figure}[tb]
\centering
\begin{tikzpicture}[scale=1]
\draw[arrows=->, >=stealth] (-0.1,0) -- (9.0,0);
\draw[arrows=->, >=stealth] (0,-0.1) -- (0,2.4);
\draw (1.0,0.0) -- (2.5,1.5);
\draw (2.5,1.5) -- (6.5,1.5);
\draw (6.5,1.5) -- (8.0,0.0);
\draw (2.5,0.2) -- (2.5,1.3);
\draw (6.5,0.2) -- (6.5,1.3);
\draw (6.5,-0.1) -- (6.5,0.1);
\draw (8.0,-0.1) -- (8.0,0.1);
\draw (-0.1,1.5) -- (0.1,1.5);
\draw (-0.1,1.5) -- (0.1,1.5);
\node at (-0.3,2.2) {$\delta_v$};
\node at (-0.3,1.5) {$\delta$};
\node[above] at (0.0,-0.55) {$0$};
\node[above] at (0.8,-0.55) {$a$};
\node[above] at (2.5,-0.56) {$a+\delta$};
\node[above] at (6.5,-0.56) {$b-\delta$};
\node[above] at (8.0,-0.55) {$b$};
\node[above] at (9.0,-0.55) {$x$};
\draw (2.5,0) circle (1.5);
\draw (2.4,-0.1) -- (2.6,0.1);
\draw (2.4,0.1) -- (2.6,-0.1);
\draw (2.1,0) circle (1.1);
\draw (2.0,-0.1) -- (2.2,0.1);
\draw (2.0,0.1) -- (2.2,-0.1);
\draw (1.375,0) circle (0.375);
\draw (1.275,-0.1) -- (1.475,0.1);
\draw (1.275,0.1) -- (1.475,-0.1);
\end{tikzpicture}
\caption{Example of a variable horizon function $\delta_v(x)$ in one dimension. The circles centered at points $x\in (a,a+\delta)$ are representations of the associated domains $H_\delta(x)$ in terms of $\delta_v(x)$. Adapted from~\cite{diehl2022coupling}.}
\label{Fig:variablehorizon}
\end{figure}
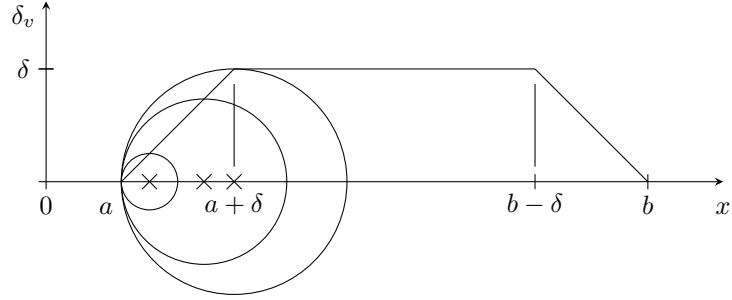

In the VHCM, the horizon $\delta_v(x)$ depends on the position $x \in \Omega$; it is usually constant in the nonlocal region far from the local-nonlocal interface and shrinks to zero when approaching the interfaces $x=a$ and $x=b$. Several choices for the transition of $\delta_v(x)$ are possible. In one dimension, it was suggested in~\cite{silling2015variable, prudhomme2020treatment, diehl2022coupling} to consider a piece-wise linear function, that is, given $\delta \in \mathbb{R}^+$ constant,
\begin{equation}
\label{eq:deltafn}
\delta_v(x) = \left\{ 
\begin{array}{ll} 
x-a, & \quad a < x \leq a+\delta, \\ 
\delta, & \quad a+\delta < x \leq b - \delta, \\ 
b - x, & \quad b-\delta < x < b,
\end{array}
\right.
\end{equation}
as shown in Figure~\ref{Fig:variablehorizon}. It follows that Equation~\eqref{eq:kappa:variable} for the stiffness constant $\kappa(x)$ now becomes:
\begin{align}
    \bar{\kappa}(x) = \frac{2EA}{\delta_v(x)^2} \text{.}
\end{align}

Using the variable horizon method, the coupling problem consists then in finding the displacements $\uelast \in \bar{\Omega}_e$ and $u \in \bar{\Omega}_\delta$ such that
\begin{equation}
\label{eq:CM-variablehorizon}
\begin{aligned}
- EA \uelast''(x) &= f_b(x), 
\quad \forall x \in \Omega_e, \\
- \int_{x-\delta_v(x)}^{x+\delta_v(x)} \bar\kappa(x) \frac{\uperid(y) - \uperid(x)}{|y-x|} dy &= f_b(x), 
\quad \forall x \in \intdomd, \\
\uelast(x) &= 0, 
\quad \text{at}\ x = 0, \\
EA \uelast'(x) &= g, 
\quad \text{at}\ x = \ell, \\
\uperid(x) - \uelast(x) &= 0, 
\quad \text{at}\ x = a, \\
\uperid(x) - \uelast(x) &= 0, 
\quad \text{at}\ x = b, \\
\sigma^+(u)(x) - EA\uelast'(x) &= 0, 
\quad \text{at}\ x=a, \\
\sigma^-(u)(x) - EA\uelast'(x) &= 0, 
\quad \text{at}\ x=b.
\end{aligned}
\end{equation}
In other words, the displacements and stresses from the two models are matched at the interfaces, i.e.\ $\uelast(x) = u(x)$ and $\sigma^\pm(x) = E\uelast'(x)$, where the peridynamic stresses $\sigma^\pm(x)$~\cite{silling2020Couplingstresses} are defined as
\begin{equation}
\label{eq:sigma2}
\sigma^\pm(u)(x)
= \frac{\delta}{2} 
\int_{x-\delta}^{x} \int_{x}^{z\pm\delta} \kappa \frac{u(y) - u(z)}{|y-z|} dydz - \frac{\kappa \delta^4}{48} u'''(x).
\end{equation}

\subsection{Discretization}
\label{sec:discretization}

Several approaches for discretizing the coupled problem~\eqref{eq:CM-variablehorizon} are available~\cite{diehl2022comparative}. Here we choose to discretize the classical linear elasticity model by a second-order finite differences scheme and the peridynamic model by a collocation approach, as introduced by Silling and Askari~\cite{Silling-Askari-CS-2005}. 

\begin{figure}[tb]
\centering
\small
\begin{tikzpicture}
\draw (0,0) -- (10.0,0);
\foreach \i in {0,...,20}
{\draw (0.5*\i,-0.1) -- (0.5*\i,0.1);}
\foreach \x in {0.0,3.5,7.5,10.0}
{\draw (\x,-0.2) -- (\x,0.1);}
\node[below] at (0.0,-0.6)  {$0$};
\node[below] at (3.5,-0.6)  {$a$};
\node[below] at (7.5,-0.6)  {$b$};
\node[below] at (10.0,-0.6) {$\ell$};
\node[below] at (0.0,-0.2)  {$x_0$};
\node[below] at (0.5,-0.2)  {$x_1$};
\node[below] at (1.0,-0.2)  {$x_2$};
\node[below] at (2.25,-0.3)  {$\ldots$};
\node[below] at (3.5,-0.2)  {$x_{n_1}$};
\node[below] at (5.5,-0.3)  {$\ldots$};
\node[below] at (7.5,-0.2)  {$x_{n_1+n_\delta}$};
\node[below] at (8.75,-0.3)  {$\ldots$};
\node[below] at (10.0,-0.2) {$x_{n}$};
\draw[arrows=<->, >=stealth] (0.0,1.75) -- (10.0,1.75);
\draw (0.0,0.25) -- (0.0,1.85);
\draw (10.0,0.25) -- (10.0,1.85);
\node[above] at (5.00,1.75) {$\Omega$};
\draw (0.0,0.45) -- (3.5,0.45);
\node[above] at (1.75,0.5) {$\Omega_1$};
\draw (3.5,0.9) -- (7.5,0.9);
\node[above] at (5.5,0.95) {$\intdomd$};
\draw (7.5,0.45) -- (10.0,0.45);
\node[above] at (8.75,0.5) {$\Omega_2$};
\foreach \x in {3.5,7.5}
{\draw (\x,0.25) -- (\x,0.90);}
\foreach \i in {0,...,7}
{\node[circle,color=black,fill=black,inner sep=0pt,minimum size=5pt,label=below:{}] at (0.5*\i,0.45) {};}
\foreach \i in {7,...,15}
{\node[circle,draw=black,fill=white,inner sep=0pt,minimum size=5pt,label=below:{}] at (0.5*\i,0.90) {};}
\foreach \i in {15,...,20}
{\node[circle,color=black,fill=black,inner sep=0pt,minimum size=5pt,label=below:{}] at (0.5*\i,0.45) {};}
\end{tikzpicture}
\caption{Definition of the grid points and degrees of freedom (represented by \(\bullet\) for the degrees of freedom associated with the classical linear elasticity model and by \(\circ\) for the degrees of freedom associated with the peridynamic model) in VHCM. Adapted from~\cite{diehl2022coupling}.}
\label{Fig:discretization}
\end{figure}
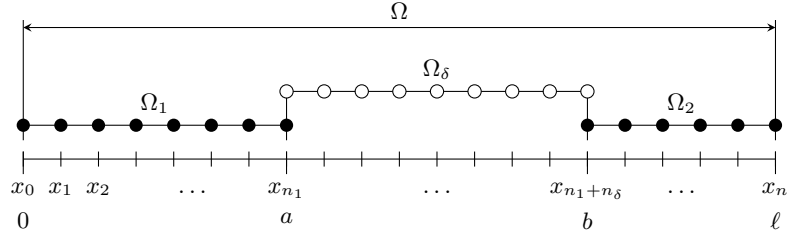

We now describe the discrete model associated with~\eqref{eq:CM-variablehorizon} on the configuration shown in Figure~\ref{Fig:discretization}. As customarily done in the literature~\cite{Silling-Askari-CS-2005,parks2008implementing}, we fix the value of the horizon $\delta$ and introduce a uniform grid spacing~$h$ chosen such that $\delta$ be a multiple of $h$, i.e.\ $\delta/h = m$, with $m$ a positive integer. Moreover, we choose the grid size $h$ to be the same in each of the subregions of the computational domain. In other words, the numbers of intervals \(n_\delta\), \(n_1\), and \(n_2\) in \(\Omega_\delta\), \(\Omega_1\), and \(\Omega_2\), respectively, are taken such that
\begin{equation}
h = \frac{b-a}{n_\delta} = \frac{a}{n_1} = \frac{\ell-b}{n_2}.
\end{equation}
Since we assume that the grid points from the two models coincide at the interfaces \(x=a\) and \(x=b\) and that the grid is uniform in the three regions \(\Omega_\delta\), \(\Omega_1\), and \(\Omega_2\), the grid points are simply given by:
\begin{equation}
x_k = k h,\quad k=0,1,\ldots,n,
\end{equation}
where $n = n_1 + n_\delta + n_2$. The numbering of the nodes is shown in Figure~\ref{Fig:discretization}. In particular, we note that $x_{n_1} = a$ and $x_{n_1+n_\delta} = b$. We consider the same numbering for the degrees of freedom associated with the displacements, that is, the nodal values will be denoted by:
\begin{equation}
u_k \approx 
\left\{
\begin{aligned}
&\uelast(x_k), &&\quad k = 0,\ldots, n_1-1\\
&\uelast(x_k)= u(x_k), &&\quad k = n_1\\
&u(x_k), &&\quad k = n_1+1,\ldots, n_1+n_\delta-1 \\
&\uelast(x_k)= u(x_k), &&\quad k = n_1+n_\delta\\
&\uelast(x_k), &&\quad k=n_1+n_\delta+1,\ldots,n
\end{aligned}
\right.
\end{equation}
where we have used the continuity of the displacement at the two interfaces, that is, $\uelast(x) = u(x)$ at $x=a$ and $x=b$. The total number of degrees of freedom is then equal to $n+1$, being each degree of freedom associated with one grid point.

The discretization of the VHC problem~\eqref{eq:CM-variablehorizon} leads to the following system of equations, in which we assume $m=2$:
\begin{enumerate}[itemsep=0pt,topsep=4pt,parsep=0pt,leftmargin=25pt]
\item 
Dirichlet BC at \(x=0\):
\begin{equation}
u_0 = 0.
\end{equation}
\item 
In \(\Omega_1\): 
\begin{equation}
\label{eq:VHCM-CLE1}
- EA \frac{u_{k-1}-2u_k+u_{k+1}}{h^2} = f_b(x_k), \quad k=1,\ldots,n_1-1.
\end{equation}
\item
At interface $x=a$:
\begin{equation}
\label{eq:couplingeqVHCMa}
EA \frac{
- 2u_{k-3} 
+ 9u_{k-2} 
-18u_{k-1} 
+11u_{k}}{6h} 
- \sigma_h^+(x_k) = 0, \quad k=n_1. 
\end{equation}
\item 
In \(\Omega_\delta\): 
\begin{equation}
\label{eq:VHCM-peridynamics}
\begin{aligned}
&- \frac{\kappa\delta^2}{2} 
\frac{u_{k-1}-2u_{k}+u_{k+1}}{h_\delta^2} = f_b(x_k),
&&\quad k=n_1+1,\\
&- \frac{\kappa\delta^2}{2} \frac{u_{k-2}+4u_{k-1}-10u_k+4u_{k+1}+u_{k+2}}{8h_\delta^2} = f_b(x_k),
&&\quad k=n_1+2,\ldots,n_1+n_\delta-2,\\
&- \frac{\kappa\delta^2}{2} 
\frac{u_{k-1}-2u_{k}+u_{k+1}}{h^2} = f_b(x_k),
&&\quad k=n_1+n_\delta-1.
\end{aligned}
\end{equation}
\item 
At interface $x=b$: 
\begin{equation}
\label{eq:couplingeqVHCMb}
EA \frac{
-11u_{k}
+18u_{k+1}
- 9u_{k+2}
+ 2u_{k+3}}{6h} 
- \sigma_h^-(x_k) = 0, \quad k=n_1+n_\delta.
\end{equation}
\item 
In \(\Omega_2\): 
\begin{equation}
- EA \frac{u_{k-1}-2u_k+u_{k+1}}{h^2} = f_b(x_k), \quad k=n_1+n_\delta +1,\ldots,n-1.
\end{equation}
\item 
Neumann BC at \(x=\ell\):
\begin{equation}
EA \frac{
- 2u_{k-3}
+ 9u_{k-2}
-18u_{k-1}
+11u_{k}}{6h} = g, \quad k=n.
\end{equation}
\end{enumerate}
We recall that if $E$ and $A$ are constant, then the quantity $\bar{\kappa}(x)\delta_v(x)^2 = \kappa\delta^2$ remains also constant. Moreover, the stresses $\sigma_h^{\pm}$ in Equations~\eqref{eq:couplingeqVHCMa} and~\eqref{eq:couplingeqVHCMb} are obtained as approximations of $\sigma_\pm$~\eqref{eq:sigma2}. More specifically, assuming that the solution $u(x)$ is sufficiently regular in $\Omega_\delta$, the stresses are approximated using one-sided third-order finite differences stencils of the first derivative as
\[
\begin{aligned}
&\sigma_{h}^+(x_{k})
= \frac{\kappa\delta^2}{2} \frac{
-11u_{k}
+18u_{k+1}
- 9u_{k+2}
+ 2u_{k+3}}{6h}, \quad k=n_1,\\
&\sigma_{h}^-(x_{k})
= \frac{\kappa\delta^2}{2} \frac{
- 2u_{k-3}
+ 9u_{k-2}
-18u_{k-1}
+11u_{k}}{6h}, \quad k=n_1+n_\delta.
\end{aligned}
\]
This choice ensures that the discretization method has a degree of precision of three, in other words, that any polynomial solution of degree up to three will be computed exactly.

\section{Methodology}
\label{sec:methodology}

We propose to employ deep learning techniques to identify the regions in which one should use the local model or the nonlocal model, by solely using load information (i.e.\ external forces), as illustrated in Figure~\ref{fig:Pipeline}. More specifically, we shall use CNNs as our deep learning tool. We describe below CNNs and the specific architecture that we considered to classify the nature of the grid points.

\subsection{Overview}

Our main objective is to develop an automated approach to identify the nonlocal region $\Omega_\delta$ and the local regions $\Omega_1$ and $\Omega_2$, shown in Figure~\ref{Fig:discretization}, to be used in a coupling method. From now on, the nodal values of the solution located in the regions $\Omega_1$ and $\Omega_2$ will be indexed by LM (Local Model), whereas the nodal values located in $\Omega_\delta$ will be indexed by NLM (Nonlocal Model). Although this is a simplified setting that does not fully reflect the numerical tests presented below, we choose to describe the method using one single nonlocal region for the sake of clarity. The extension to more than one nonlocal domains is straightforward.

\begin{figure}[tb]
\centering
\includegraphics[width=1\textwidth]{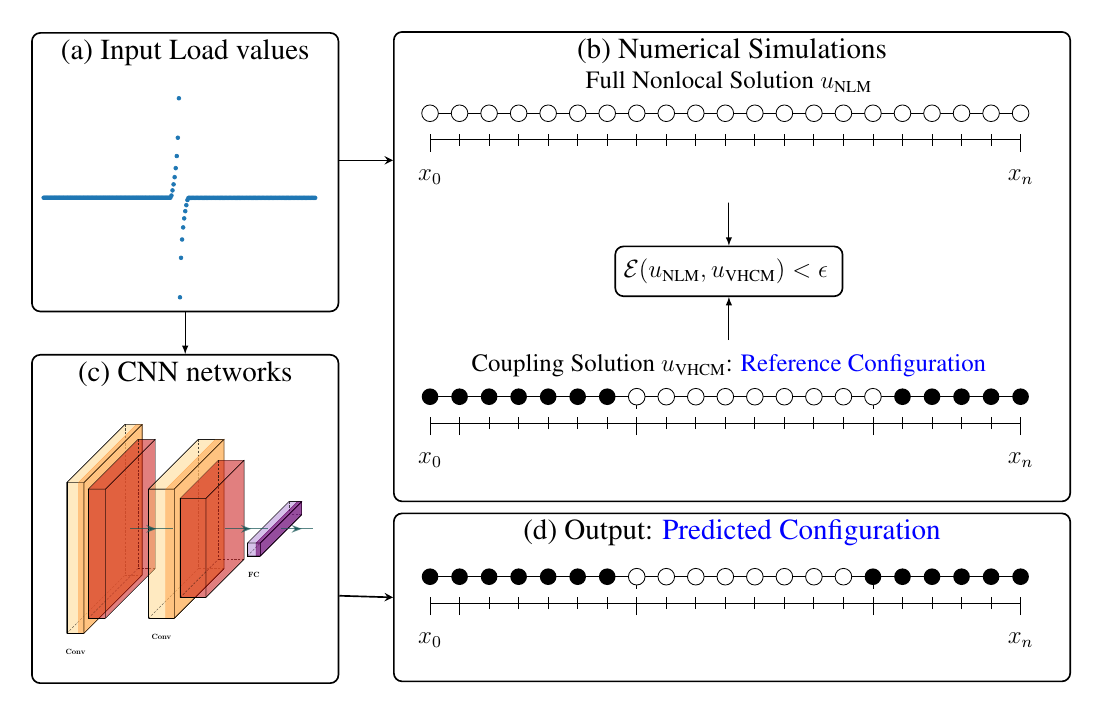}
\caption{Overview of the training and prediction: (a) Input Load values used for data generation and as network inputs; (b) Numerical simulations: fully nonlocal solution and coupled solution obtained with the reference local-nonlocal domain splitting; (c) CNN network model; (d) Network output: predicted local-nonlocal splitting configuration. Filled circles \(\bullet\) indicate local nodes and empty circles \(\circ\) indicate nonlocal nodes. }
\label{fig:Pipeline}
\end{figure}

An overview of the proposed automated identification process is shown in Figure~\ref{fig:Pipeline}. The input data consists of the external load vector $f_b(x_k)$, $k=1,\ldots,n-1$, evaluated at all interior grid points~$x_k$, see Figure~\ref{fig:Pipeline}{a}. The output of the process is a label associated with each node that takes the value~0, if the node is classified as belonging to the local region, or~1 otherwise (i.e.\ the node belongs to the nonlocal region). It is essential to note that the nature of our approach is supervised learning, that is, each load input during training corresponds to a known classification. As this requires training data, the load input (see Figure~\ref{fig:Pipeline}{a}) is also used to generate the ground-truth data that will serve as the reference configuration during training. The generation, as shown in Figure~\ref{fig:Pipeline}{b}, is conducted using the following procedure: a coupling configuration (local-nonlocal domain assignment) is added to the training set when it provides a coupled solution whose error, with respect to the fully nonlocal solution, is below a given tolerance. More specifically, we first compute the reference displacement $u_\text{NLM}$ using the nonlocal model in which all nodal points are NLM nodes~\(\circ\). Then, given a coupling configuration (in which the nodes are either LM nodes~\(\bullet\) or NLM nodes~\(\circ\)), we compute the displacement $u_\text{VHCM}$ using the method described in the previous section, see Figure~\ref{fig:Pipeline}{b}.
As mentioned above, the reference configuration is included in the training dataset if the relative error in the $\ell_2$ norm is below a given tolerance $\varepsilon$, namely
\begin{equation}
\label{eq:l2_norm}
\mathcal E(u_\text{NLM},u_\text{VHCM}) =\frac{\lVert u_\text{NLM} - u_\text{VHCM}\rVert_{\ell_2}}{\lVert u_\text{NLM} \rVert_{\ell_2}} < \varepsilon.
\end{equation}
For more details about the data generation, the reader is referred to Section~\ref{sec:data_generation}. The loads and the reference configurations of the local-nonlocal splitting provide the training data for the CNN in a supervised manner. Once trained, the CNN model takes as input an (unseen) external load $f_b(x)$, see Figure~\ref{fig:Pipeline}c, and outputs the predicted configuration (i.e.\ it classifies the nodes as either LM \(\bullet\) or NLM \(\circ\) at each grid point) in the domain, see Figure~\ref{fig:Pipeline}d. More details of the CNN architecture are given in the following section.

\subsection{Convolution neural networks}
\label{sec:CNN_section}

In recent years, deep learning has yielded impressive results across a spectrum of challenges, such as in visual recognition, speech understanding, natural language processing, and even inference of physical systems responses. Among different deep neural network architectures, CNNs have been investigated and analyzed in depth~\cite{gu2018recent}. CNNs are particularly effective for tasks involving data with a grid-like topology, such as images, audio spectrograms, and sequential data arising for instance in natural language processing,  with applications ranging from image classification and object tracking to speech and natural language processing~\cite{gu2018recent, li2021survey}.

The CNN structure consists of various fundamental components, including convolutional, pooling, and fully connected layers. The standard structure involves a series of convolution layers followed by a pooling layer. The layers are then succeeded by one or more fully connected layers. The phase in which input data undergoes transformation into output through the layers is the so-called forward propagation process~\cite{yamashita2018convolutional}. In the next two sections, we give a brief description of CNNs' features for the non-expert reader.

\begin{enumerate}[itemsep=4pt,topsep=4pt,parsep=0pt,leftmargin=25pt]

\item
\textbf{Convolutional Layers:} 
Convolutional layers are the cornerstone of CNNs. These layers use learnable kernels that convolve over the input data, enabling the network to detect local patterns and features. More precisely, a kernel is typically a small array, usually of size 3~$\times$~3, which is systematically passed through an input array, also referred to as a tensor. At each position within the tensor, an element-wise multiplication is performed between the elements of the kernel and the corresponding elements in the input tensor. A bias is then added to the sum of these products in order to yield the output value, which occupies the corresponding position in the output tensor—termed ``feature map''. This process is reiterated by employing multiple kernels to generate diverse feature maps. These feature maps encapsulate various attributes of the input tensors, such as distinct kernel functions serving as distinct feature extractors. The operation of a convolutional layer can be expressed as:
\begin{equation}
\label{eq:convolution_layer}
A_j = \sigma_\text{ReLU}
\bigg(\sum_{i=1}^{N} C_i K_{i,j} + B_j \bigg),
\end{equation}
where $C_i$ denotes an input matrix, $K_{i,j}$ a kernel matrix,  $N$ is the number of feature maps from the previous layer, and $B_j$ a bias value. Finally, the rectified linear unit activation function $\sigma_\text{ReLU}$~\cite{sharma2017activation},
\[
\sigma_\text{ReLU}(x)=\max(0,x),
\]
is applied element-wise to the matrix, resulting in the output matrix $A_j$.
The primary task of the learning process involves identifying sets of suitable kernel matrices capable of extracting distinctive and discriminative features for the output tasks. To achieve this, the back-propagation algorithm, commonly employed to optimize the connection weights in neural networks, can be adapted. In this context, it serves to train both the kernel matrices and biases, functioning as shared neuron connection weights.

\item
\textbf{Pooling Layers:}
Pooling layers downsample the dimensions of feature maps while retaining the salient information; thus, their main effect is an intelligent feature-dimension reduction. To effectively reduce the number of output neurons, pooling algorithms usually combine neighboring elements in the convolution output matrices. 
Notable among these algorithms are max-pooling and average-pooling~\cite{albawi2017understanding}.  In this study, we implement a max-pooling layer with a  2 × 1 size, which selects the highest value from the two neighboring elements of the input feature map, thereby generating a single element for the output feature map. 


\item
\textbf{Fully Connected Layers:}
Fully connected layers process the flattened feature vectors to make the final predictions. A fully connected layer operation with input $z$ and output $y_{\text{out}}$ is represented as
\begin{equation}
y_{\text{out}} = \sigma(Gz + v)
\label{eq:fully_connected_layer}
\end{equation}
where $y_{\text{out}}$ is the output, $z$ the input vector, $G$ the weight matrix, $v$ the bias vector, and $\sigma$ a given activation function~\cite{sharma2017activation}.
\end{enumerate}

\begin{figure}[tb]
\centerline{
\includegraphics[width=1\textwidth]{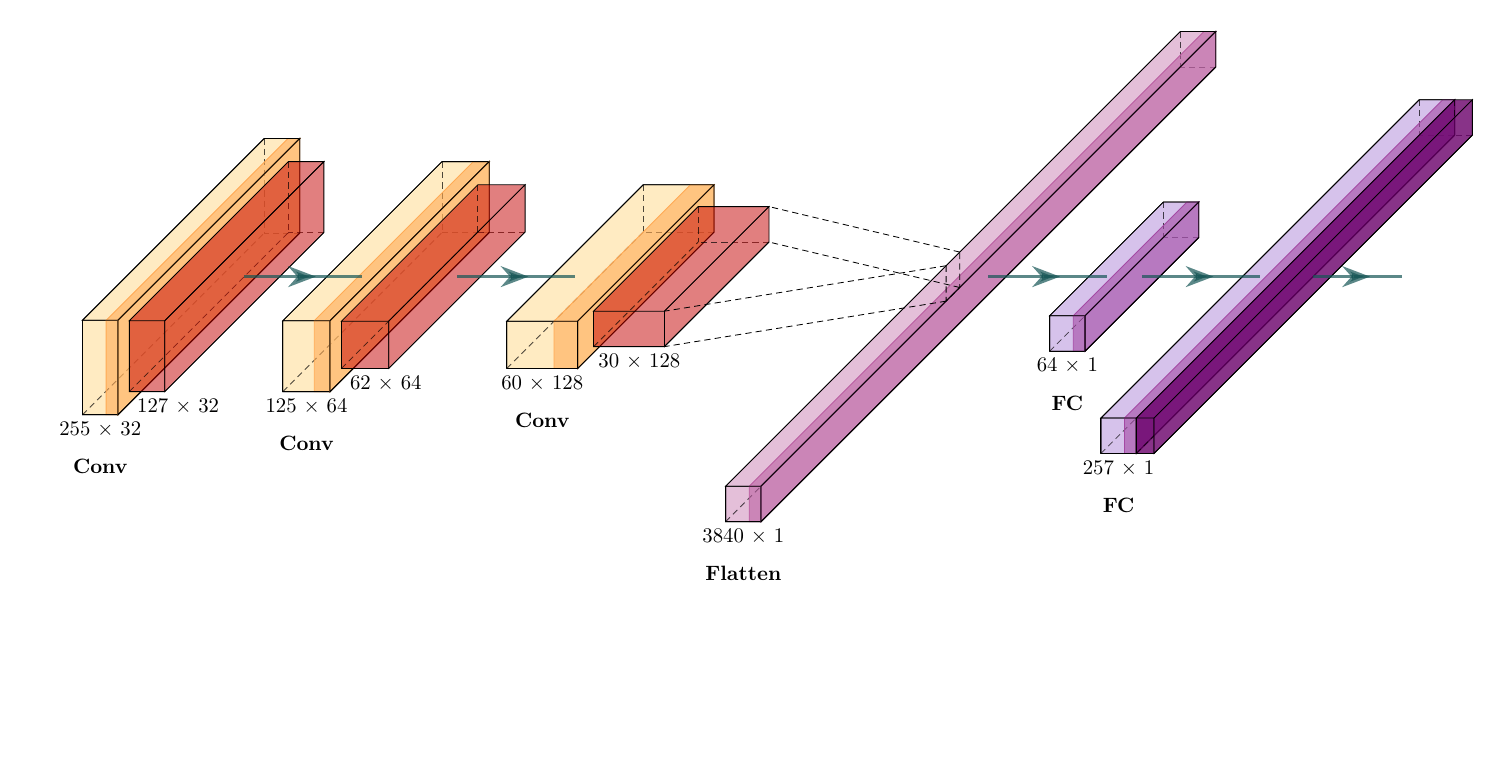}}
\caption{Architecture of the proposed CNN model.}
\label{fig:Cnn}
\end{figure}

\paragraph{The selected CNN architecture:} We propose to use the following CNN architecture for the identification of local and nonlocal regions. First, as a preprocessing step, the input of the network is normalized to a load vector with unit variance and zero mean. The architecture is composed of convolutional layers with ReLU activation functions for feature extraction, followed by pooling layers for spatial reduction, and fully connected layers for the final output. This is shown in Fig.~\ref{fig:Cnn}. The first convolution layer uses 32 filters with $3 \times 1$ kernels. The second convolution layer has 64 filters with 3 kernels, and the third convolution layer has 128 filters with $3\times 1$ kernels. Each convolution layer is followed by a max pooling layer with pool size of two. The output tensor from these three conv-pooling blocks has a dimension of $30 \times 128$. The following two layers consist of one flattened layer, followed by a fully connected layer with 64 neurons. The last layer of our CNN model is a fully connected (dense) layer, where the number of neurons matches the output size. In our study, we explore two cases with distinct output sizes. In the first case, the CNN receives as input the full vector of the second derivative of the load functions at the grid points. The resulting output is a vector of labels, having the same size of the input load vector. Herein, the output size is 257, leading to a fully connected layer with 257 neurons. 
Meanwhile, in the second case, the CNN receives a window as input. The resulting output is a label for the central node. Herein, the output size is then one, corresponding to a single neuron in the final layer. Further details on these cases are discussed in Section \ref{sec:case_studies}. The sigmoid activation function is used in the last layer across both cases. It is important to note that our experimental investigations explore a variety of network architectures, using different numbers of layers and neurons. Out of all the architectures evaluated, the one that produces the best results is the one that is presented herein.

\paragraph{Computational resources and model parameters:}
The CNN model is trained by using the Adam optimizer \cite{Kingma2014} and a batch size of 32. The training proceeds for a maximum of 200 epochs; however, an early stopping callback is implemented to avoid overfitting. When early stopping is used, training will end as soon as the validation dataset's loss does not improve \cite{bisong2019regularization}. The learning rate is set to the commonly used value of 0.001. The proposed model is implemented in TensorFlow 2.10, with
Keras in Python 3.10.11. 






\subsection{Classification evaluation metrics}
\label{sec:evaluation}


In the context of binary classification problems such as ours, the classification capability of the trained network is usually evaluated with a confusion matrix.
The confusion matrix displays the correctly or incorrectly predicted results. It is based on four key elements: true positives (TP, correctly predicted LM), false positives (FP, wrongly predicted LM), true negatives (TN, accurately predicted NLM), and false negatives (FN, wrongly predicted NLM). To evaluate the confusion matrix, we use two induced metrics, namely the accuracy metric and the F1-score, which are the most used metrics in classification~\cite{hossin2015review,lu2021review}:

\begin{enumerate}[itemsep=4pt,topsep=4pt,parsep=0pt,leftmargin=25pt]
\item
The accuracy metric measures the
ratio of correct predictions over the total
number of instances evaluated. Higher values for accuracy are indicative of better performance. Formally,
\begin{align}  
\text{Accuracy} = \frac{\text{TP} + \text{TN}}{\text{TP+FP+TN+FN}}.
\end{align}



\item
The F1-score is the harmonic mean of ``precision'' and ``recall''. It ranges from 0 to 1, where a higher value indicates better classification performance. Formally,
\begin{align}
    \text{F1-score} = 2 \times \frac{\text{Precision} \times \text{Recall}}{\text{Precision} + \text{Recall}},
\end{align}
where 
\[
\begin{aligned}
& \text{Precision} = \frac{\text{TP}}{\text{TP} + \text{FP}}, \\
& \text{Recall} = \frac{\text{TP}}{\text{TP} + \text{FN}}.
\end{aligned}
\]
\end{enumerate}

We note that in {\bfseries case 1}, see Section~\ref{sec:numerical:results:fulldomain}, we report the average over all samples for all metrics (i.e.\ confusion matrix, accuracy, and F1-score).

\section{Data generation}
\label{sec:data_generation}

The CNN used for classification is trained on input-output pairs of external loads and corresponding coupling configurations (inducing a coupled solution with a certain accuracy); this pipeline is illustrated in Figure~\ref{fig:Pipeline}. In this section, we introduce the definition of the load functions and present the steps involved in the generation of the coupling regions.


\subsection{Families of loading functions}
\label{sec:loading_functions}

In this section, we consider several classes of loading functions, including singular functions that induce discontinuous solutions with a finite jump to mimic fracture phenomena in higher dimensions. One advantage of peridynamics is that the model does not feature any spatial derivatives thus allowing for solutions with jumps. 

\begin{figure}[tb!]
\centering
\begin{minipage}{0.35\textwidth}
\subfloat[]{\includegraphics[height=0.15\paperheight]{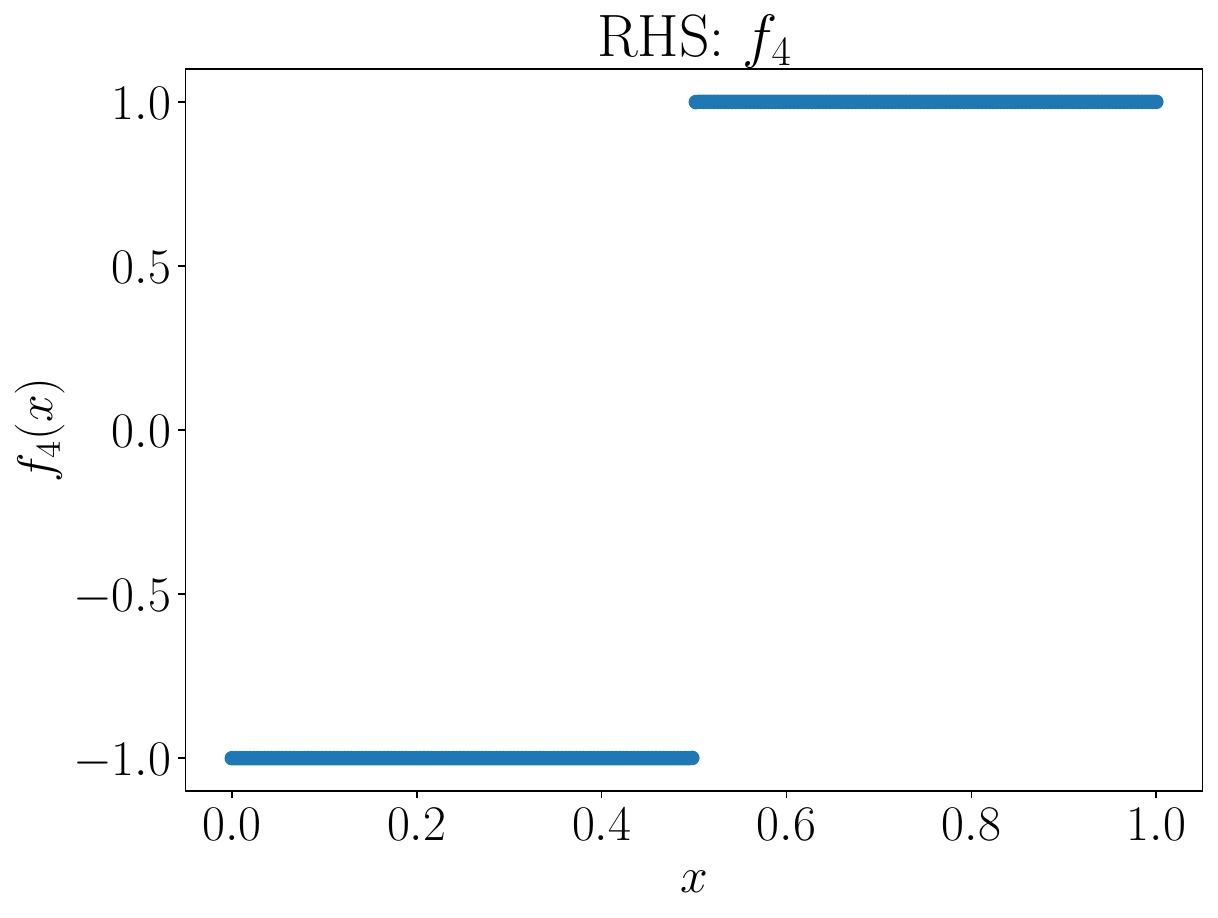}}
\par
\subfloat[]{
\includegraphics[height=0.15\paperheight]{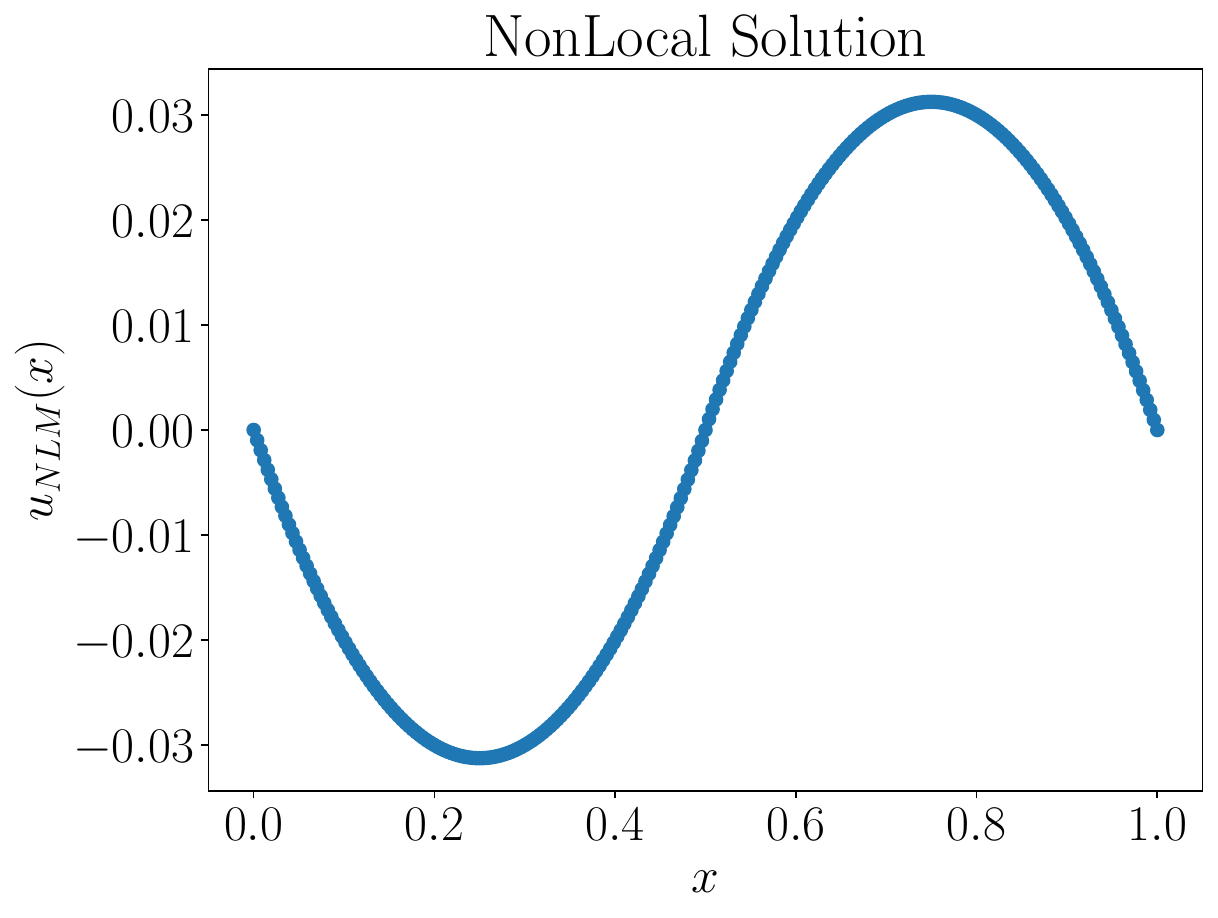}}
\end{minipage}   
\begin{minipage}{0.35\textwidth}
\subfloat[]{%
\includegraphics[height=0.15\paperheight]{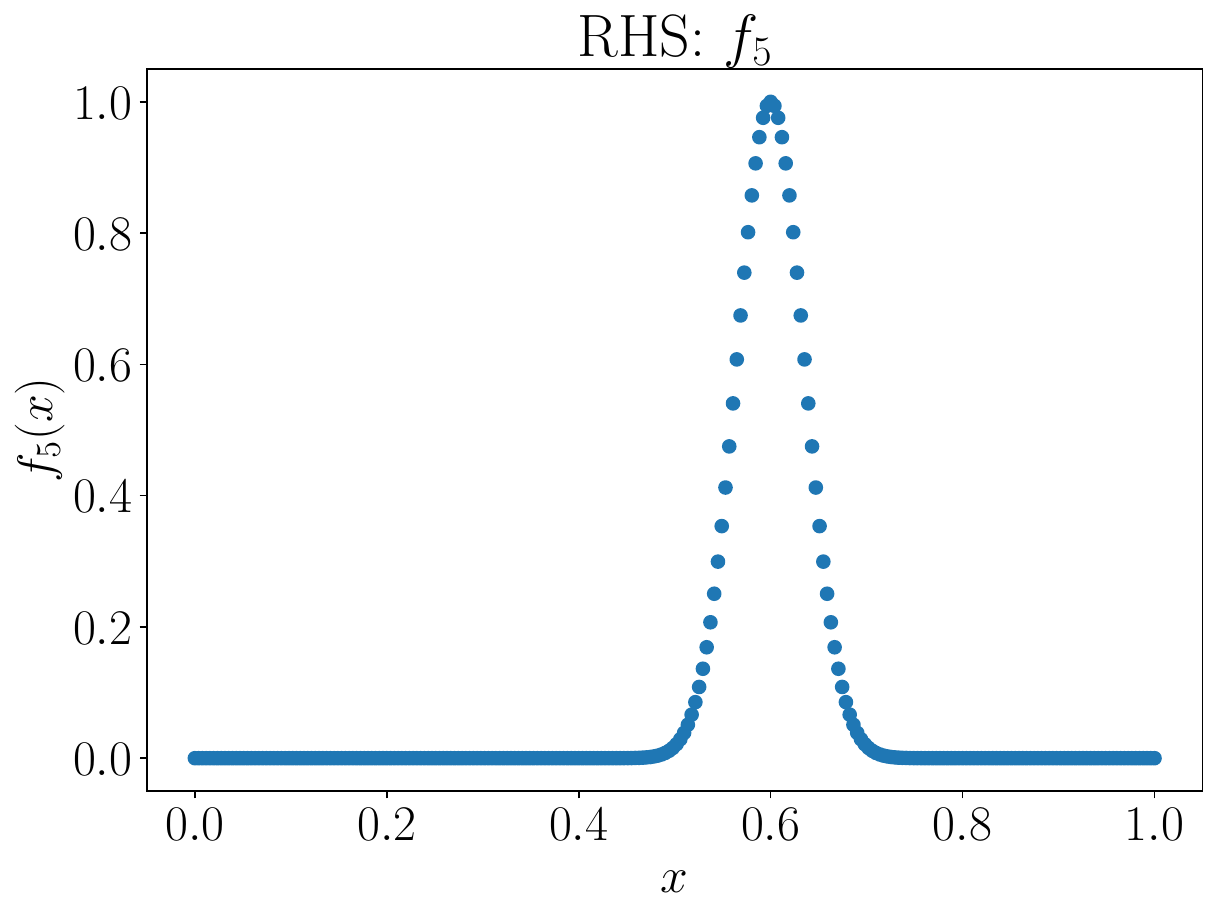}}
\par
\subfloat[]{%
\includegraphics[height=0.15\paperheight]{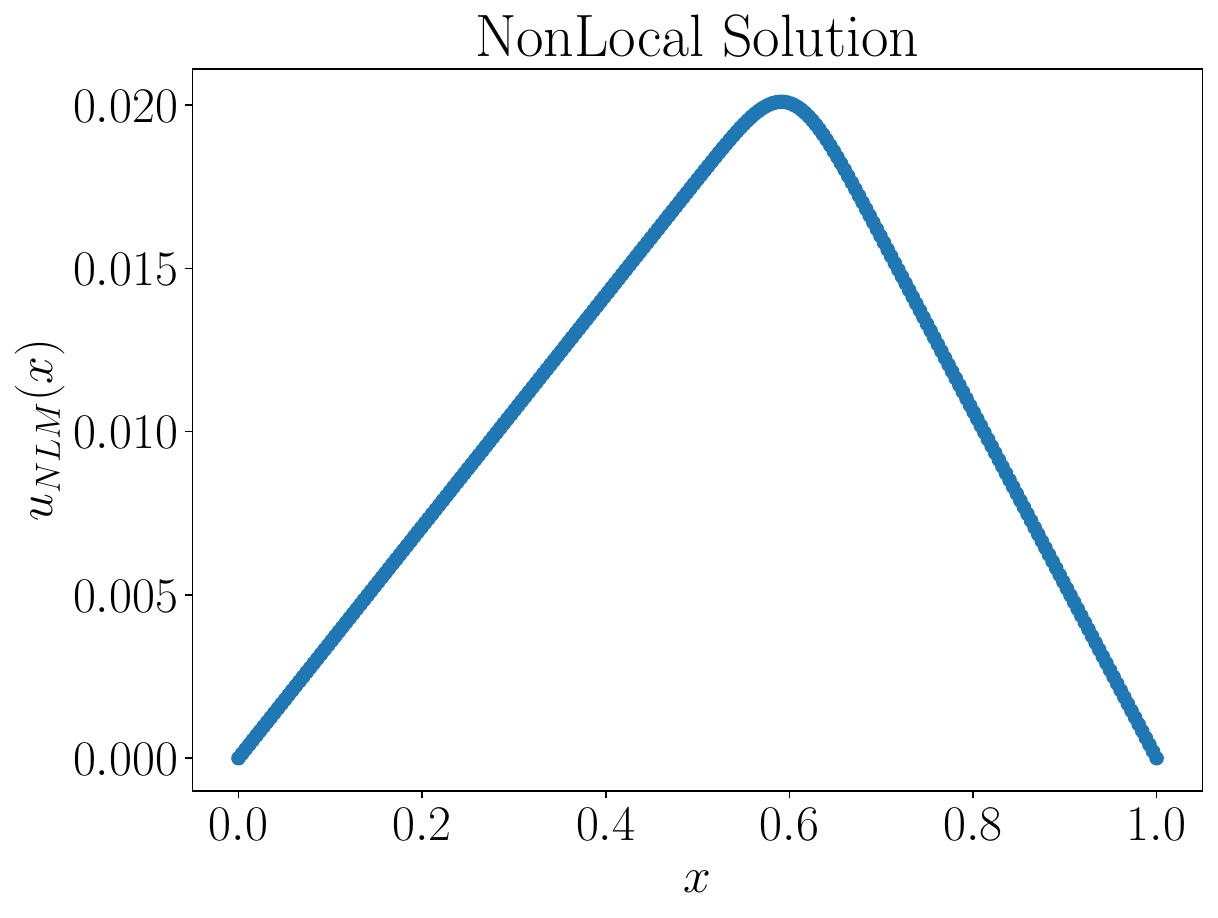}}
\end{minipage}   
\caption{RHS functions $f_i(x)$, $i=4$, $5$ used as test functions and their associated nonlocal solution. (a) Forcing function $f_4$ in Eq.~\eqref{eq:force_f4} for $t = 0.0005$; (b) Corresponding full nonlocal solution $u_{\text{NLM}}$; (c) Forcing function $f_5$ in Eq.~\eqref{eq:force_f5} for $c = 0.6$; (d) Corresponding full nonlocal solution $u_{\text{NLM}}$.}
\label{fig:f4f5}
\end{figure}

\begin{enumerate}[itemsep=4pt,topsep=4pt,parsep=0pt,leftmargin=25pt]
\item

            
The first family of load functions, $f_1$, is based on functions~\cite{chen2011continuous} featuring a singularity at $x = 0.5$ and defined as:
\begin{equation}
\label{eq:f1}
f_1(x)=
\begin{cases}
0, & \mbox{for}\ x\in [0, 0.5  - \delta),\\
\frac{1}{2}\delta^2 -\delta +\frac{3}{8} + (2\delta - \frac{3}{2}  -\ln{\delta})x \\
\quad\quad + (\frac{3}{2} +\ln{\delta}) x^2 -(x^2 -x)\ln({(\frac{1}{2}-x)}), & \mbox{for}\ x\in [0.5-\delta, 0.5),\\
\frac{1}{2}\delta^2 -\delta +\frac{3}{8} + (2\delta + \frac{3}{2}  +\ln{\delta})x \\
\quad\quad - (\frac{3}{2} +\ln{\delta}) x^2 +(x^2-x) \ln({x-\frac{1}{2}}), & \mbox{for}\ x\in [0.5, 0.5+\delta),\\
1,  & \mbox{for}\ x\in [0.5+\delta, 1.0].
\end{cases}
\end{equation}

The function $f_1$ and the corresponding fully nonlocal solution $u_\text{NLM}$ are shown in Figure~\ref{fig:all}(a-b) for $\delta$ = 0.03 and $h=\delta/8$. 

In cases where the singularity is located at a point $x_\text{jump}$ different from $0.5$, we will consider the loading function $f_1(x+(0.5 -x_\text{jump}))$.

\begin{figure}[tb!]
\centering
\begin{minipage}{0.325\textwidth}
\subfloat[]{
\includegraphics[height=0.14\paperheight] {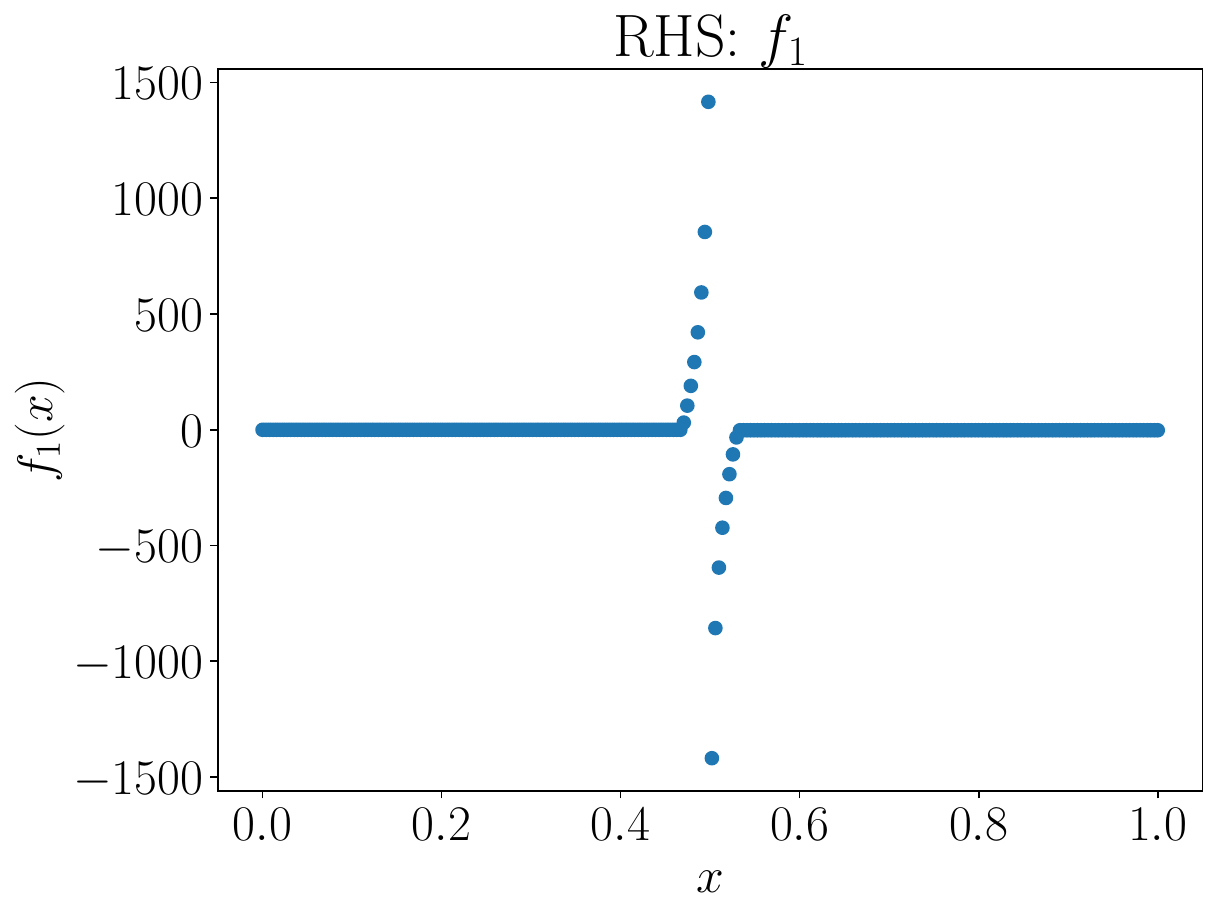}}
\par
\subfloat[]{
\includegraphics[height=0.14\paperheight]{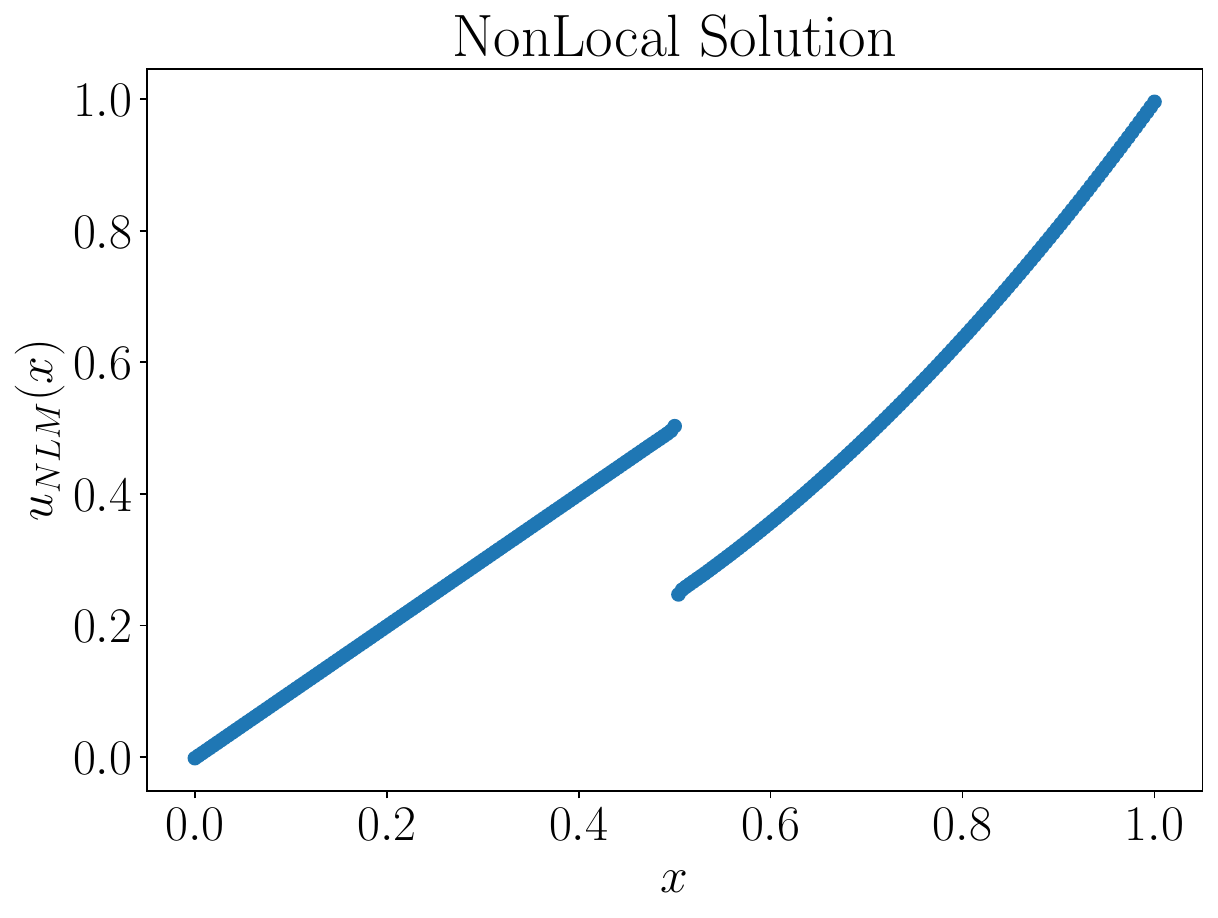}}
\end{minipage}
\begin{minipage}{0.325\textwidth}
\subfloat[]{
\includegraphics[height=0.14\paperheight]{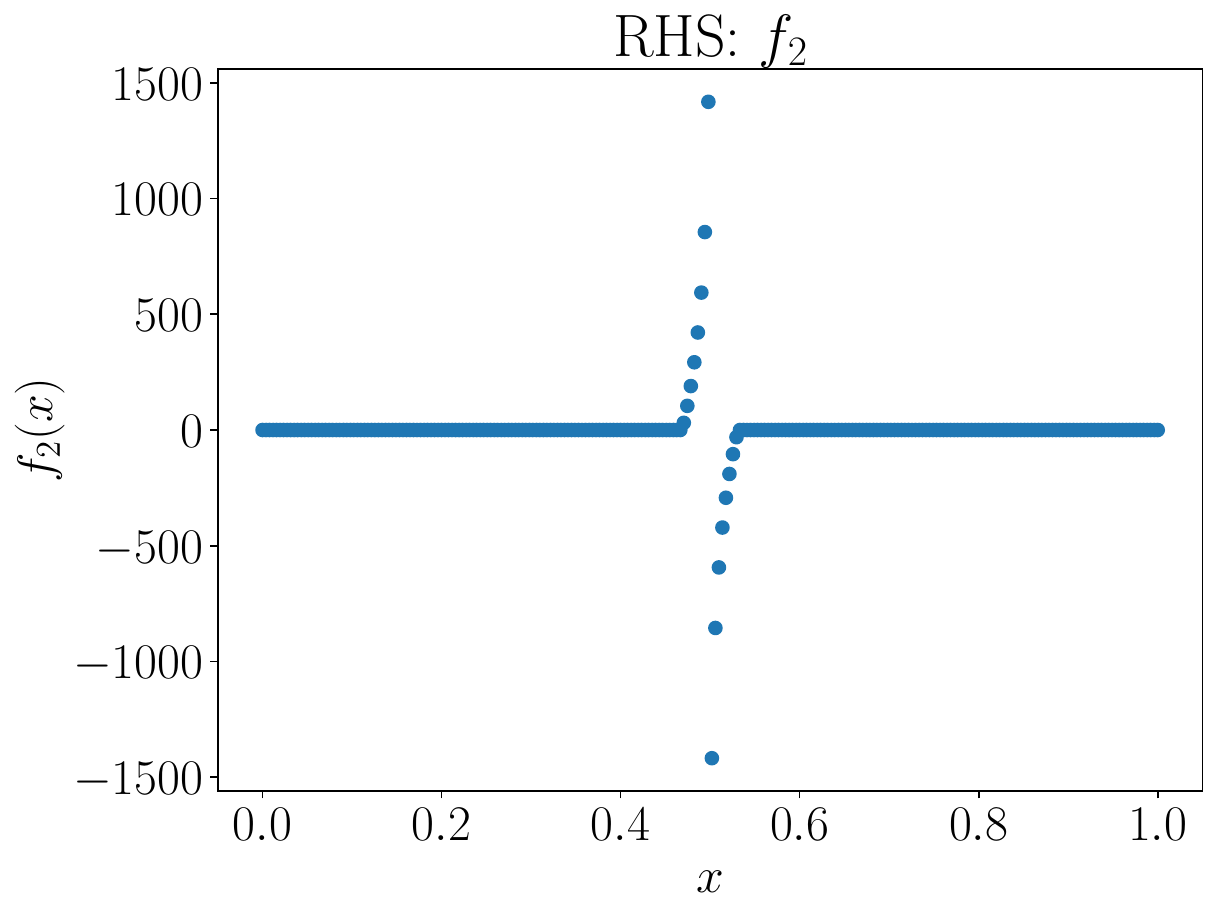}}
\par
\subfloat[]{
\includegraphics[height=0.14\paperheight]{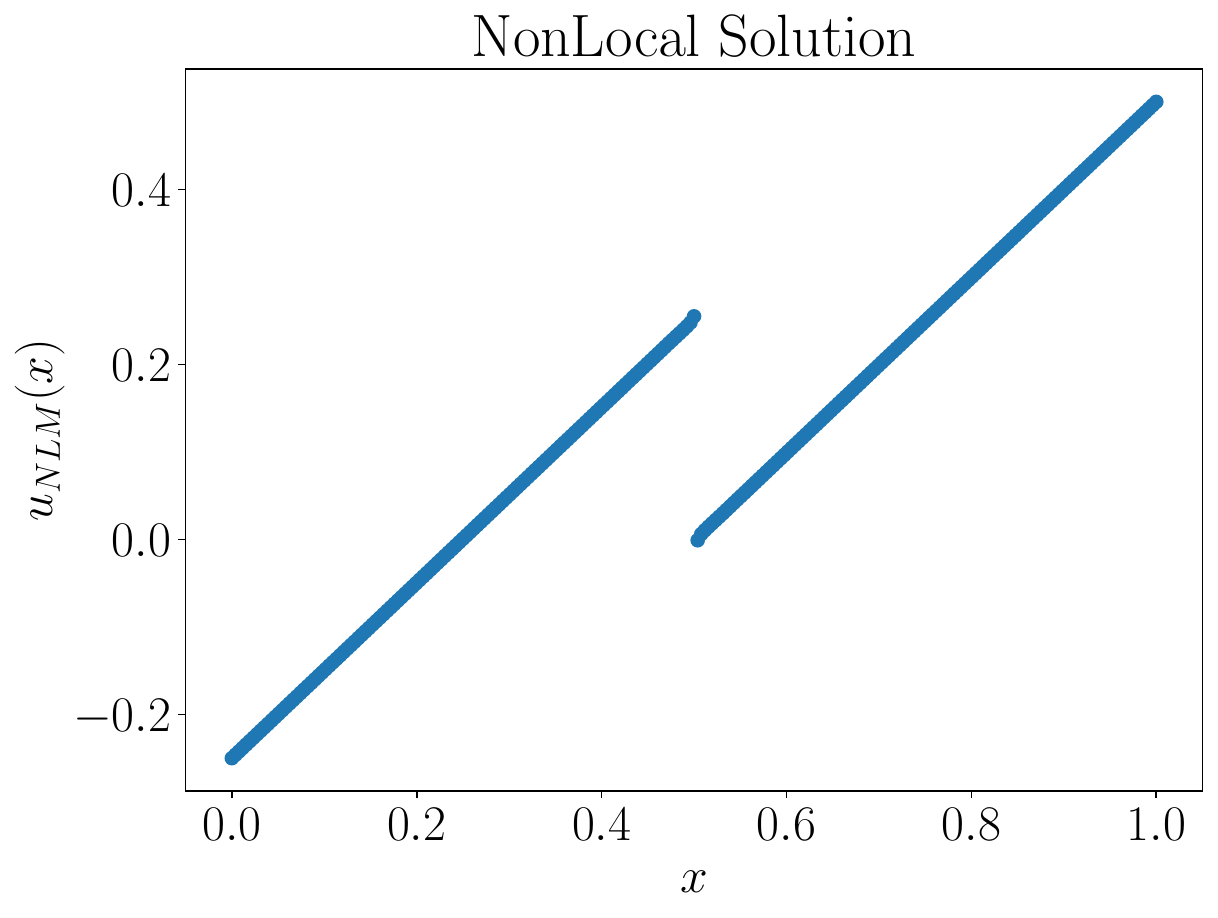}}
\end{minipage}
\begin{minipage}{0.325\textwidth}
\subfloat[]{
\includegraphics[height=0.14\paperheight]{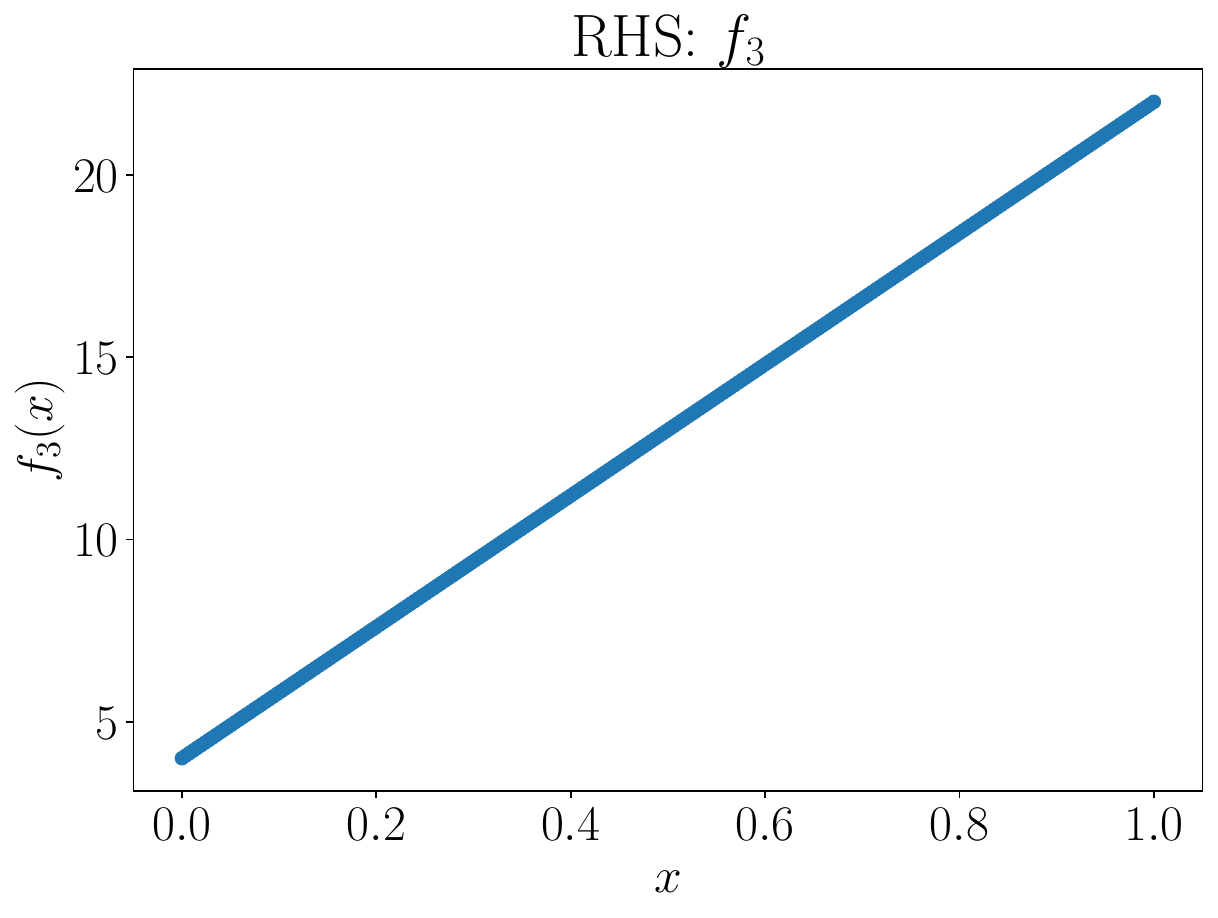}}
\par
\subfloat[]{
\includegraphics[height=0.14\paperheight]{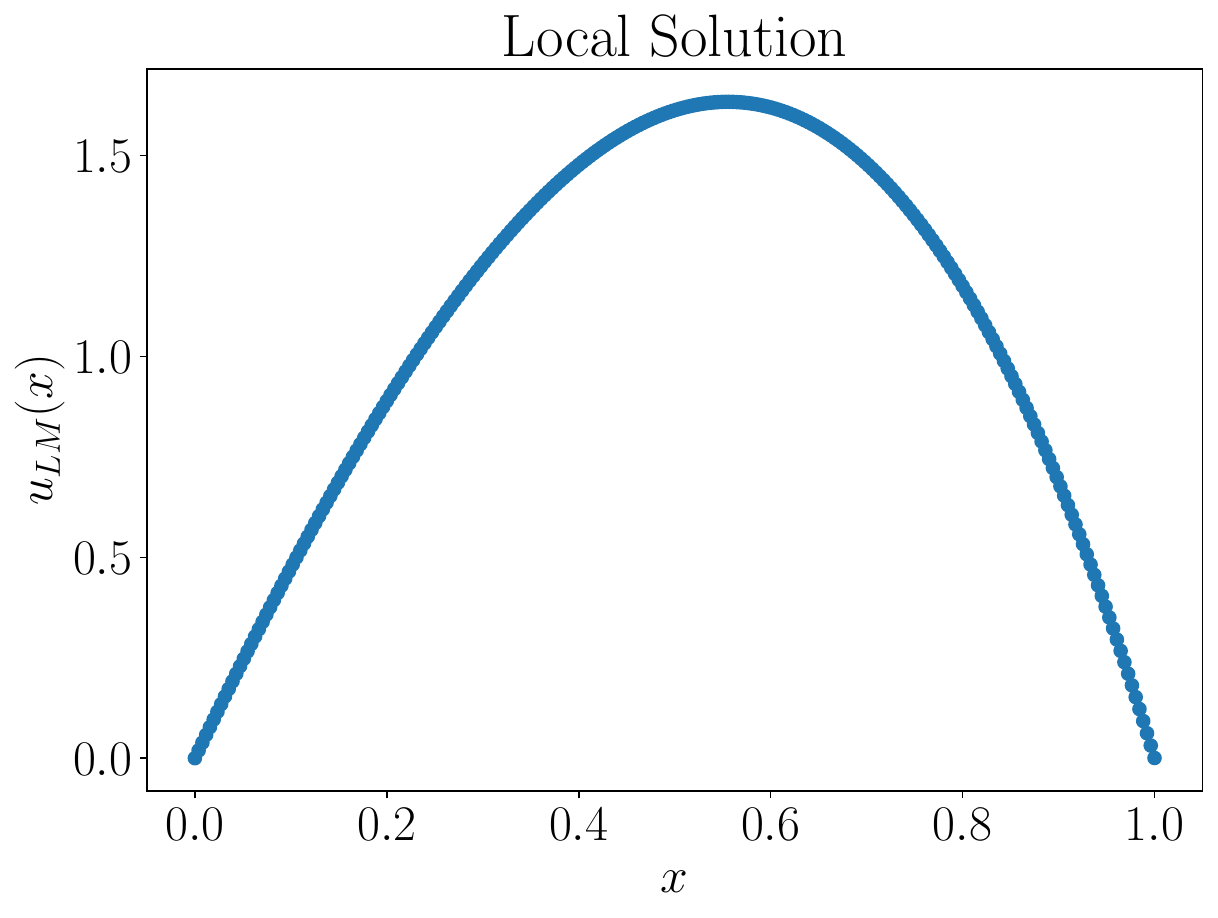}}
\end{minipage}
\caption{RHS functions $f_i(x)$, $i=1$, $2$, and $3$ with and without discontinuity used during training and their associated nonlocal/local solution. (a) Forcing function $f_1$ in Eq.~\eqref{eq:f1}; (b)  Corresponding full nonlocal solution $u_{\text{NLM}}$; (c) Forcing function $f_2$ in Eq.~\eqref{eq:f2}; (d) Corresponding full nonlocal solution $u_\text{NLM}$; (e) Forcing function $f_3$ in Eq.~\eqref{eq:force_f3} for $c_1=18$ and $c_2=4$; (f) Corresponding full local solution $u_{\text{LM}}$.}
\label{fig:all}
\end{figure}

            

\item
The second family of load functions, $f_2$, featuring a singularity at $x = 0.5$~\cite{d2014optimal}, is defined by:
\begin{equation}
\label{eq:f2}
f_2(x)=\begin{cases}
0, 
& \mbox{for}\ x\in [0, 0.5 - \delta),\\
\frac{1}{2\delta^2} \big(\ln{\delta}-\ln({\frac{1}{2}-x}) \big), 
& \mbox{for}\ x\in [0.5-\delta, 0.5),\\
\frac{1}{2\delta^2} \big(\ln({x-\frac{1}{2}})-\ln{\delta} )\big),
& \mbox{for}\ x\in [0.5, 0.5+\delta),\\
0,  
& \mbox{for}\ x\in [0.5+\delta, 1.0].
\end{cases}
\end{equation}

The function $f_2$ and the corresponding full nonlocal solution $u_\text{NLM}$ are plotted in Figure~\ref{fig:all}(c-d) for $\delta$ = 0.03 and $h=\delta/8$. 

In cases where the singularity is located at a point $x_\text{jump}$ different from $0.5$, we will evaluate the loading function as $f_2(x+(0.5 -x_\text{jump}))$.

\item
The study also includes the family of loads characterized by polynomial solutions of degree three and lower. The loading functions \(f_3(x)\) are in this case linear of the form:
\begin{equation}
\label{eq:force_f3}
f_3(x)=c_1x + c_2,
\end{equation}
an example of which is shown in Fig.~\ref{fig:all}(e-f) with $c_1=18$ and $c_2=4$.


\item
We consider as well the following family of load functions:
\begin{equation}
\label{eq:force_f4}
f_4(x) = \tanh\frac{1}{t} \bigg(x - \frac{1}{2} \bigg),
\end{equation}
where the parameter $t$ controls the variation in the solution around the point $x=0.5$. An example of function $f_4$ is shown in Fig.~\ref{fig:f4f5}(a-b) for $t=0.0005$.
\item
Finally, we also include a family of smooth load functions that induce continuous solutions with local behavior: 
\begin{equation}
\label{eq:force_f5}
f_5(x) = e^{-400(x-c)^2}.
\end{equation}
Such a function is a smooth representation of a point-wise load applied at the point $x=c$. The load and corresponding solution is shown in Fig.~\ref{fig:f4f5}(c-d) for $c=0.6$.
\end{enumerate}

\subsection{Generation of the coupling regions}
\label{sec:ref_config}

We present the steps used to generate the output data for the CNN training, i.e. we describe how the reference configurations of the coupling regions are constructed based on the loading functions.

For the sake of simplicity, all experiments focus on a one-dimensional bar with a material property of $E=1$ and cross-sectional area $A=1$, resulting in $\kappa = 2/\delta^2$. We consider the configuration of Figure~\ref{Fig:discretization} with $\ell= 1$ and interface locations $a$ and $b$. We choose the values of $a$ and $b$ such that grid points coincide the two interfaces. The domain is partitioned into sub-intervals of size $h = 1/n$, with $n$ given. The horizon is taken as a multiple of $h$, i.e.~$\delta=mh$, with $m=8$. 
We acknowledge that implementing non-uniform grids in our approach poses significant challenges, particularly in integrating them with the standard CNN architecture and our coupling method. Therefore, and for the purposes of this proof-of-concept, we will employ a fixed discretization grid to evaluate the innovative aspects of our proposed approach. 
To avoid any issues with applying the local boundary conditions, we assume that $\Omega_\delta = (a, b)$ is such that $a > 2\delta$ and $b < \ell-2\delta$. We will nevertheless vary the positions $a$ and $b$ of the interfaces in order to consider solutions with a discontinuity located in the interval $(2\delta,\ell-2\delta)$. 


We now describe the process of generating the reference coupling regions used in the training for the special cases $f_1$ and $f_2$.
Since these loads are smooth away from their respective singularities, it seems sensible that the region where one would use the nonlocal model in the local-nonlocal coupling approach should contain the singular point.
We thus compute the fully nonlocal reference solution $u_\text{NLM}$ and choose the smallest interval around the singularity, i.e.\ \(\Omega_{\delta}=(x_\text{jump}-\alpha,x_\text{jump}+\alpha)\) where \(\alpha>0\) is a multiple of the mesh size $h$, such that the relative error $\mathcal E(u_\text{NLM},u_\text{VHCM})$~\eqref{eq:l2_norm} between the fully nonlocal and coupled solutions is below a given tolerance \(\epsilon\). In our experiments, we fix \(\epsilon\) to \(0.01\).
It is clear that the proposed process is directly influenced by the specific features of the chosen loading functions. 

\begin{figure}[tb!] 
\centering
\subfloat[]{%
    \includegraphics[width=0.5\textwidth]{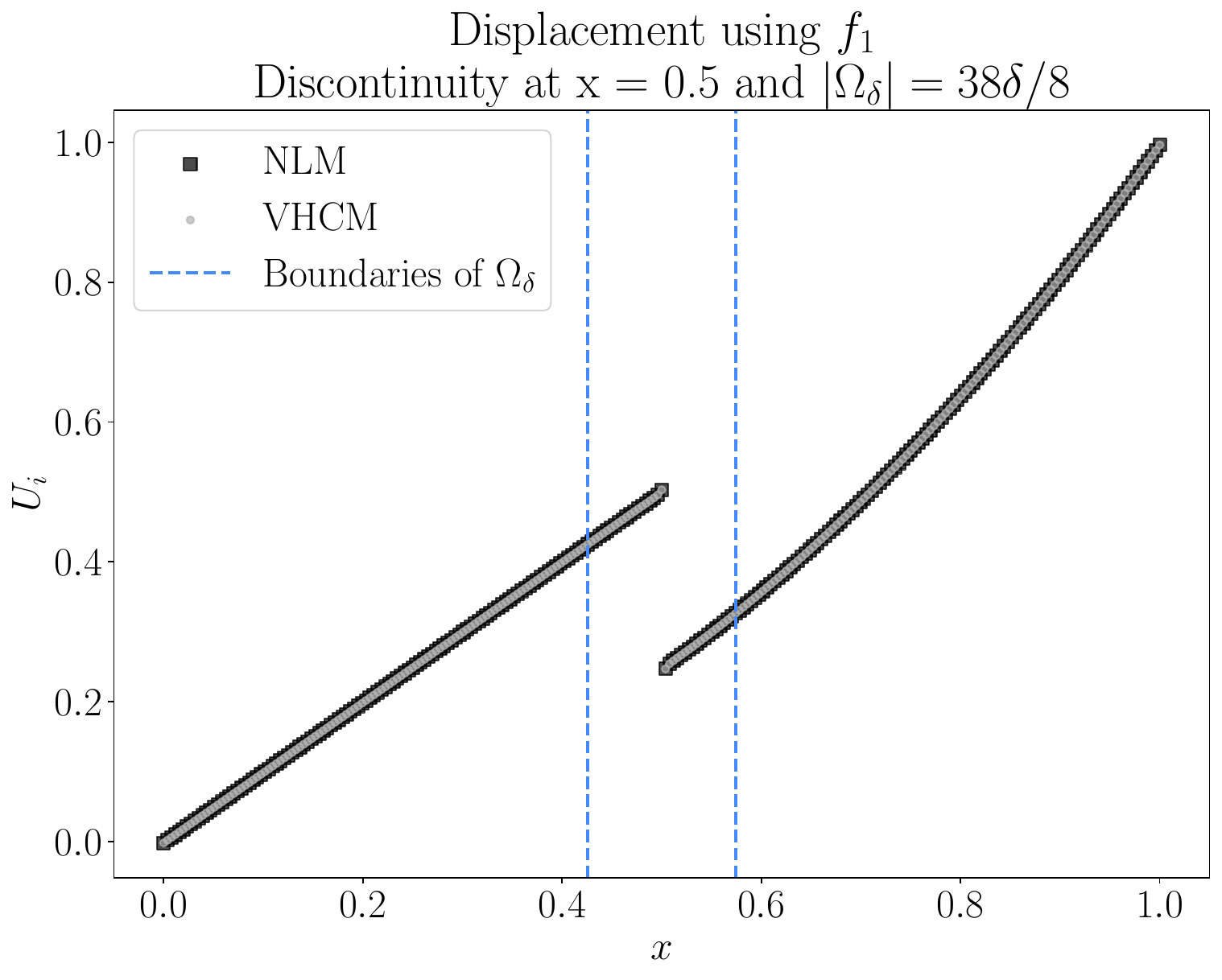}}
\hfill
\subfloat[]{%
    \includegraphics[width=0.5\textwidth]{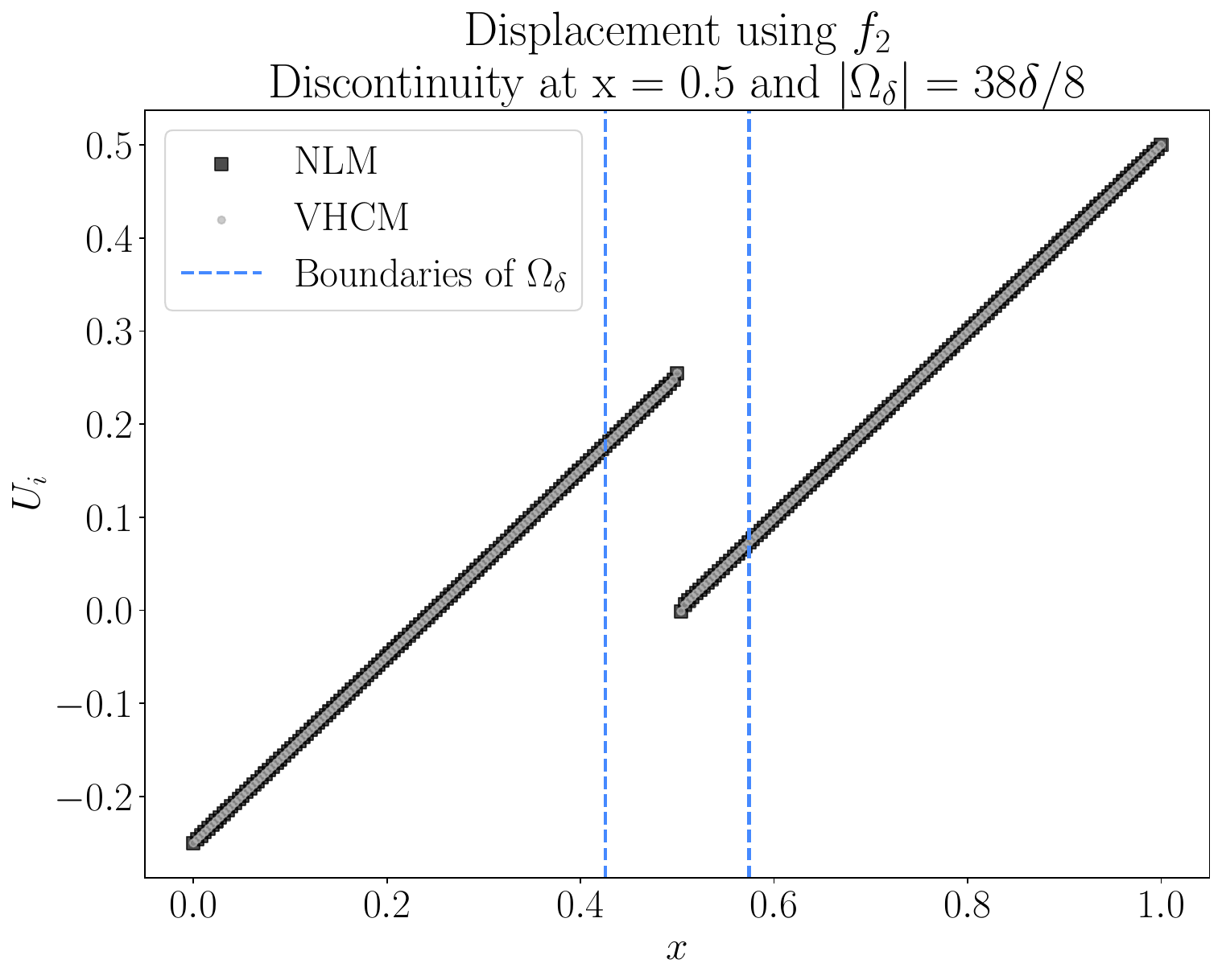}}
\vfill
\subfloat[]{%
    \includegraphics[width=0.5\textwidth]{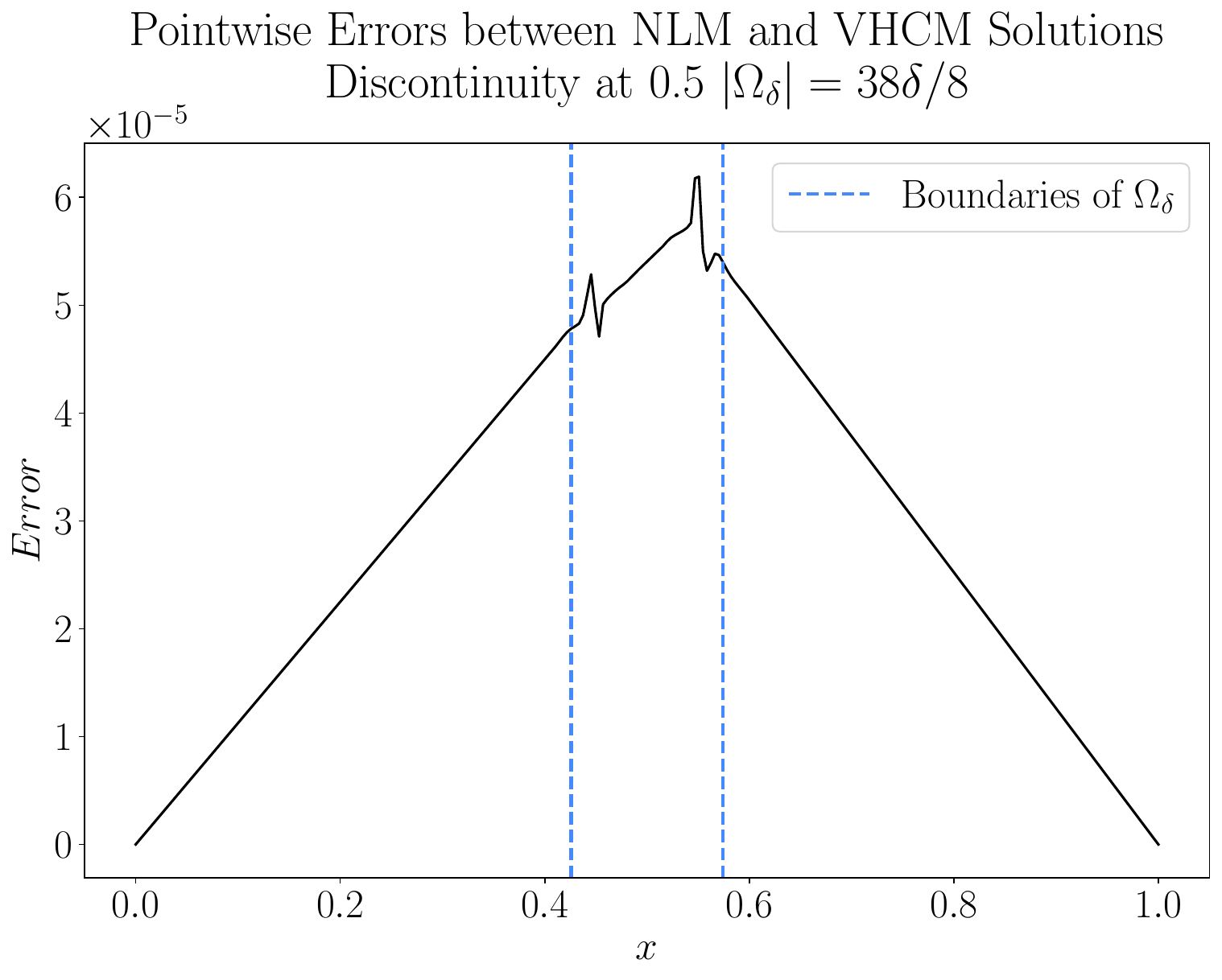}}
\hfill
\subfloat[]{%
    \includegraphics[width=0.5\textwidth]{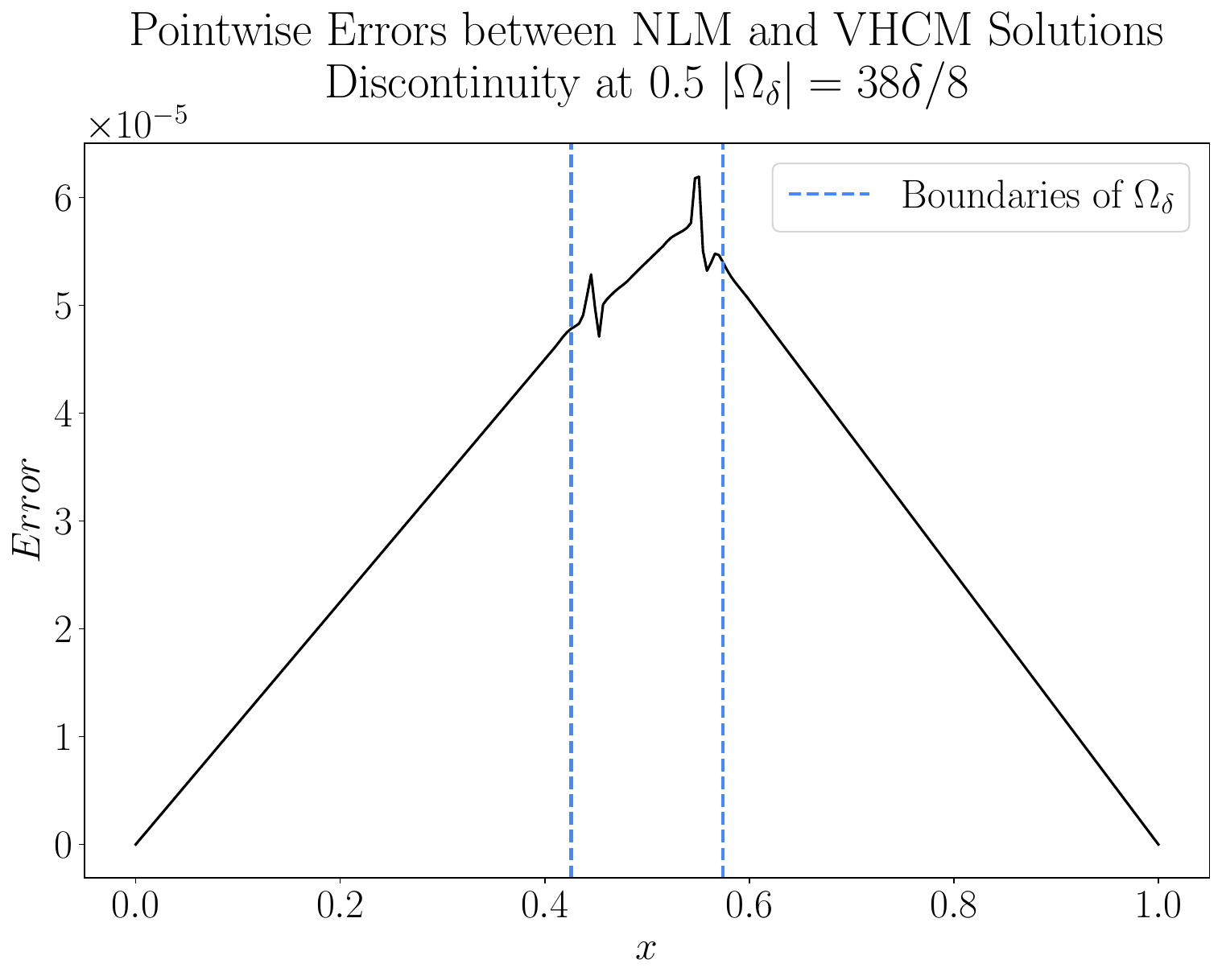}}
\caption{Peridynamic region configuration with discontinuity at $x=0.5$, and corresponding pointwise errors between the solutions.  $u_{\text{NLM}}$ is represented by \textcolor{black}{$\blacksquare$} and $u_{\text{VHCM}}$ is represented by~\textcolor{gray}{$\bullet$}. (a)~Displacement fields $u_\text{NLM}$ and $u_\text{VHCM}$ obtained with $f_1$ and $\mathcal E(u_\text{NLM},u_\text{VHCM})=7.026\times 10^{-3}$; (b)~Displacement fields $u_\text{NLM}$ and $u_\text{VHCM}$ obtained with $f_2$ and $\mathcal E(u_\text{NLM},u_\text{VHCM})=7.027\times 10^{-3}$. Note that the two curves superpose in both cases. Pointwise errors between $u_\text{NLM}$ and $u_\text{VHCM}$ using (c) $f_1$ and (d) $f_2$.}
\label{fig:uc_upd}
\end{figure}

In Figure~\ref{fig:uc_upd}, we show an example of an optimal nonlocal region configuration using $f_1$ and $f_2$ with singularity at $x = 0.5$. The displacements $u_\text{NLM}$ and $u_\text{VHCM}$ using $f_1$ are shown in Figure~\ref{fig:uc_upd}(a). The value of $\mathcal E(u_\text{NLM},u_\text{VHCM})$ using $f_1$ is $7.026 \times 10^{-3}$.  Figure~\ref{fig:uc_upd}(b) presents the displacement $u_{\text{NLM}}$ and $u_{\text{VHCM}}$ using the loading type $f_2$. The value of $\mathcal E(u_\text{NLM},u_\text{VHCM})$ in this case is $7.027 \times 10^{-3}$. In both cases, the error in the coupled solution is below the selected threshold. We also show the pointwise error between the nonlocal solution (NLM) and the coupled solution (VHCM) in Figure~\ref{fig:uc_upd}({c-d}), to illustrate that the local error remains relatively small at all points.

As mentioned in Section~\ref{sec:models}, the solutions to the local and the nonlocal models coincide for polynomials of degree up to three; furthermore, it is known that the fourth derivative of the solution is an indicator of the extent of the nonlocality of the solution \cite{chen2011continuous}. As a pre-processing step, we leverage this information by taking the second derivative of the loading functions rather than the original functions themselves. Here, the second derivatives are simply evaluated using finite differences approximations of the loading functions. This step will help in accurately capturing the underlying behavior for the load and serve as a nonlocality indicator. The second derivatives of these loads are then used as inputs for our ML model.

We now proceed to explain the process when using load $f_3$. The coefficients \(c_1\) and \(c_2\) in Eq.~\eqref{eq:force_f3} are random values within the range \([-30,30]\) and \([-10,10]\), respectively.
In this case, the reference configuration is  characterized by a fully local region as the local model and the nonlocal model lead to the same solution. This  choice is made to anchor the training of our model on an instance where the behavior is fully local. This is pivotal for enriching the learning process of the model, enabling it to discern and accurately predict the nuances of fully local configurations alongside the more complex local-nonlocal configurations. Notably, the second derivative of $f_3$ is always zero and is used as input for our ML model.

\subsection{Approaches for input-output selection}
We recall that we consider two strategies. In the first one, we utilize the complete dataset over the full computational domain as input. In this scenario, the nature of the problem is akin to the simultaneous classification of all nodes. In the second approach, we transform the datasets into windowed inputs, and solve a point-wise classification problem. 

\subsubsection{Full-domain input data}
\label{sec:whole_input}

The first strategy consists of considering the second derivative of the load functions at all grid points simultaneously following the generation of data as explained in Section~\ref{sec:ref_config}. In this context, we increase the model's data volume and enhance its ability to generalize by introducing specific transformations to the functions $f_1$~\eqref{eq:f1} and $f_2$~\eqref{eq:f2}. Specifically, we incorporate the negation operation, which complements our existing data by providing contrasting cases.
The second derivatives of $f_1$, $f_2$, and $f_3$ are estimated and used here as input vectors.
In this approach, the input-output pairs are given by a full second derivative of the load vector-full labels vector. The CNN receives the full second derivative  vector as input. The resulting output is a vector of labels, having the same size of the input vector. Each output node has either the \text{LM} or \text{NLM} label, according to the region (local or nonlocal) they belong to. 
After prediction, we implement a post-processing step where nonlocal regions containing fewer than eight nodes are treated as local (LM) since we are using the variable horizon coupling method (VHCM) with $m=8$.

\begin{table}[tb!]
    \setlength\tabcolsep{6pt} 
\centering 
\caption{Overview of dataset distribution and associated processing times.}
\label{tab:table1}
\resizebox{0.8\textwidth}{!}{%
\begin{tabular}{lcc}
    \toprule
       \multicolumn{3}{c}{\textbf{A. Dataset Distribution}} \\
   
    \midrule
  & \textbf{Case 1: Full-domain Data} & \textbf{Case 2: Windowing Data} \\
    \midrule
    Train       & 589 & 22802 \\
    Test        & 119  & 4498  \\
    Validation  & 78  & 3476 \\
    Total       & 786 & 30776 \\
    \midrule
    \multicolumn{3}{c}{\textbf{B. Processing Time (seconds)}} \\
    \midrule
      & \textbf{Case 1: Full-domain Data} & \textbf{Case 2: Windowing Data} \\
    \midrule
    Generation Time & 138.77 & 182.37 \\
    Train Time      & 30.19 & 137.17 \\
    Test Time       & 0.6    & 10.9 \\
    \bottomrule
\end{tabular}%
}
\end{table}

\subsubsection{Window-based input data}
\label{sec:window_sec}

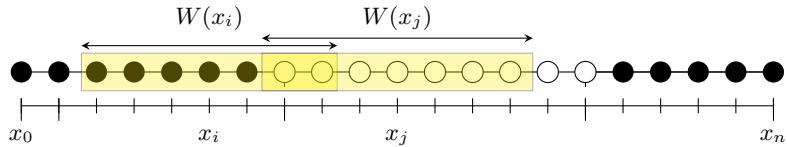
\begin{figure}[tb!]
\centering
\small
\begin{tikzpicture}

\draw (0,0) -- (10.0,0);
\foreach \i in {0,...,20}
{\draw (0.5*\i,-0.1) -- (0.5*\i,0.1);}
\foreach \x in {0.0,0.5,3.5,7.5,10.0}
{\draw (\x,-0.2) -- (\x,0.1);}

\node[below] at (0.0,-0.2)  {$x_0$};
\node[below] at (2.5,-0.2)  {$x_i$};
\node[above] at (2.5,0.9)  {$W(x_i)$};
\draw[arrows=<->, >=stealth] (0.8,0.8) -- (4.2,0.8);

\node[below] at (5,-0.2)  {$x_j$};
\node[above] at (5,0.9)  {$W(x_j)$};
\draw[arrows=<->, >=stealth] (3.2,0.9) -- (6.8,0.9);

\node[below] at (10.0,-0.2) {$x_n$};

\draw (0.0,0.45) -- (10.0,0.45);

\draw (7.5,0.45) -- (10.0,0.45);

\foreach \x in {3.5,7.5}
{\draw (\x,0.25) -- (\x,0.55);}

\foreach \i in {0,...,7}
{\node[circle,color=black,fill=black,inner sep=0pt,minimum size=8pt,label=below:{}] at (0.5*\i,0.45) {};}

\foreach \i in {7,...,15}
{\node[circle,draw=black,fill=white,inner sep=0pt,minimum size=8pt,label=below:{}] at (0.5*\i,0.45) {};}

\foreach \i in {16,...,20}
{\node[circle,color=black,fill=black,inner sep=0pt,minimum size=8pt,label=below:{}] at (0.5*\i,0.45) {};}

\draw[fill=yellow, opacity=0.3] (0.8,0.2) rectangle (4.2,0.7);

\draw[fill=yellow, opacity=0.3] (3.2,0.2) rectangle (6.8,0.7);
\end{tikzpicture}
\caption{Schematic representation of data points and their corresponding windows. Two specific data points, $x_i$ and $x_j$, are highlighted, each associated with its own window, denoted as $W(x_i)$ and $W(x_j)$.}
\label{fig:windowing_approach}
\end{figure}

The second (and more effective) strategy proposed in this paper is the windowing approach. This approach proves to be a highly effective strategy for handling data with several singularities. 
In this approach, each node, denoted as $x_i$, is treated individually.
In the windowing approach, we reformulate the classification problem as point-wise classification; an illustration of the window selection is presented in Figure~\ref{fig:windowing_approach}. By utilizing window configurations around each data point, we are effectively segmenting the load's second derivative, enabling a focused and precise classification strategy at each point. This method allows for a more focused classification approach, providing a detailed insight into the load characteristics within specific windowed segments. The induced input-output pairs are then given by window-label. The CNN receives a window of load's second derivative as input. The resulting output is a label for the central node.  The label is either the \text{LM} or \text{NLM}, according to the region (local or nonlocal) the central node belong to. In a post-processing step, we apply the same process detailed in Section~\ref{sec:whole_input}.
This approach aims at enhancing the model's capability to detect patterns and variations in the data, making our analysis more precise and adaptable to different load scenarios.

We present in Figure~\ref{fig:windowing_approach} an illustration of the window selection for the window-based approach. Considering two points and their corresponding windows, the window spans two horizons on the left and on the right of the central point, thereby covering a total of four horizons. Figure~\ref{fig:windowing_approach} shows two central nodes $x_i$ and $x_j$, each with its associated windows $W(x_i)$ and $W(x_j)$. These windows are used as input data for the CNN enabling, classification labels based on their region. The output labels will be then \text{LM} for $x_i$ since it is located in the local region (represented by $\bullet$) and \text{NLM} for $x_j$ since it is located in the nonlocal region (represented by \(\circ\)). 

We note that this approach can be also applied and used in higher dimensions. For instance, in two or three dimensions, the windows might transition from one-dimensional segments to two-dimensional areas or even three-dimensional volumes. By varying these windows, we aim to capture and understand how nonlocal effects act within specific regions or volumes across the broader dimensional space.

\subsection{Case studies}
\label{sec:case_studies}
In this section, we describe the different scenarios and case studies considered in our tests. A summary is reported in Table~\ref{tab:case_studies}.

\subsubsection{Case 1: Full-domain input data}
\label{sec:one:discontinuity}

In this case, we generate the data as explained in Section~\ref{sec:whole_input}. These simulations result in a total of 3,084 data samples, as shown in Table~\ref{tab:table1}(A). The data is first split into training, testing, and validation sets. The training set is used to train the CNN model, where the model learns the relationship between inputs and outputs, adjusting its parameters (like weights and biases) accordingly. The validation set is used during the training process; it helps in evaluating the performance of the model on a dataset it has not be subjected to before. The testing set is used to check how well the model can generalize and helps prevent overfitting. In other words, the testing set allows one to evaluate the performance of the model on new, unseen data~\cite{raschka2018model}.  We split the dataset by assigning 75\% of the samples to the training set, 10\% to the validation set and 15\% to the testing
set. The size of each data sample is 257, resulting in a training set of size $589\times 257$ and a validation set of size $78\times 257$, while the testing set has size $119\times 257$. The generation and partitioning of the dataset into training, validation, and testing sets necessitates a total of 138.77 seconds (Table~\ref{tab:table1}(B)).

In the full-domain input data approach, the CNN receives as input the full vector of the second derivatives of the load functions at the grid points. The resulting output is a vector of labels, having the same size of the input load vector. Each output node has either the \text{LM} or \text{NLM} label, according to the region (local or nonlocal) they belong to.
The CNN model is structured as multiple node classification model; the results are detailed in Section~\ref{sec:numerical:results:fulldomain}. A summary of this test case is shown in Table~\ref{tab:case_studies}.

\begin{table}[tb]
\center
    \setlength\tabcolsep{6pt} 
        \renewcommand{\arraystretch}{1.5}

    \caption{Summary for training case studies.}
    \begin{tabular}{
        |>{\raggedright\arraybackslash}m{2cm}
        |>{\raggedright\arraybackslash}m{5.6cm}
        |>{\raggedright\arraybackslash}m{5.8cm}|}
        \hline
        
        \textbf{Case Study} & \textbf{Case 1} & \textbf{Case 2}
        \tabularnewline[0.3em]
        \hline

        \textbf{Section} & Section \ref{sec:numerical:results:fulldomain} & Section \ref{sec:numerical:results:classification}  
        \tabularnewline[0.1em]
        \hline
            
        \textbf{Input}  &Full Load's Second Derivative Vector & Windows 
        \tabularnewline[0.1em]
        \hline
        
        \textbf{Output} &Full Labels Vector  & One Node Label
        \tabularnewline[0.1em]
        \hline
        
        
        \textbf{Load Dataset} 
        & 
        $f_i(x)$, $i=1,2,3$. 
        &
        $f_i(x)$, $i=1,2,3$.
        \tabularnewline[0.2em]
        \hline
        
        \textbf{ML Model Type} & Multiple node Classification & \multicolumn{1}{l|}{Node wise Classification}
        \tabularnewline[0.2em]
        \hline 
        \textbf{Interpretation} & 
       
        \parbox{5.5cm}{
        Demonstrated the feasibility of the proposed approach for region detection, as a kind of a ``proof of concept".
        } & 
        \parbox{5.8cm}{
        \vspace {2mm} 
        Highly effective strategy for handling data with varying numbers of discontinuities without the need for retraining the model for each specific scenario.
        \vspace {2mm} }
        \tabularnewline[0.2em]
        \hline
    \end{tabular}
    \label{tab:case_studies}
\end{table}

\subsubsection{Case 2: Window-based input data}
\label{sec:Windowing}

In this case, we use the window-based approach, in which the dataset is generated and preprocessed using the methodology explained in Section~\ref{sec:window_sec}. The initial dataset, consisting of 786 complete input-output vector pairs, is partitioned into distinct sets for training, validation, and testing. The training set constitutes 75\% of the total data, the validation set allocates 10\%, and the remaining 15\% is designated for the testing set. We then implement the windowing approach, segmenting each vector into windows of data. Subsequently, we perform a deduplication process, removing duplicated input windows-output label  in each set and between sets (train-test, train-validation, and validation-test), ensuring unique samples across all sets.
The resultant windowed dataset comprises a total of 30776 data samples (Table~\ref{tab:table1}(A)). The training set, the validation set, and the testing set contain 22802, 3476, and 4498  data samples, respectively. We note that the size of input samples depends on the window size.
The generation and data splitting into windows require a total of 182.37 seconds (Table~\ref{tab:table1}(B)). We point out that we introduced the deduplication step to ensure fairness in the testing process. While straightforward in  1D experiments with uniform grids, in practice and especially in higher dimensions, deduplication is nontrivial. However, it is often the case that solutions have common features in certain regions of a computational domain so that having duplicates is very natural and actually helps the learning process. Thus, our results show that even when we stress test the model by removing favorable data, the model still performs well (Section \ref{sec:numerical:results:classification}).

The CNN inputs are the windows and the corresponding output is the label (local or nonlocal) of the central node. Thus, the model has been restructured as a node-wise classification model. We recall that the labels are either \textbf{LM} or \textbf{NLM} depending on the classification (local or nonlocal). An overview of this case is summarized in Table~\ref{tab:case_studies}.

\section{Numerical results}
\label{sec:numerical:results}

\subsection{Case 1: Full-domain input data} 
\label{sec:numerical:results:fulldomain}



\begin{figure}[tb!] 
\centering
    \subfloat[]{%
        \includegraphics[width=0.56\textwidth]{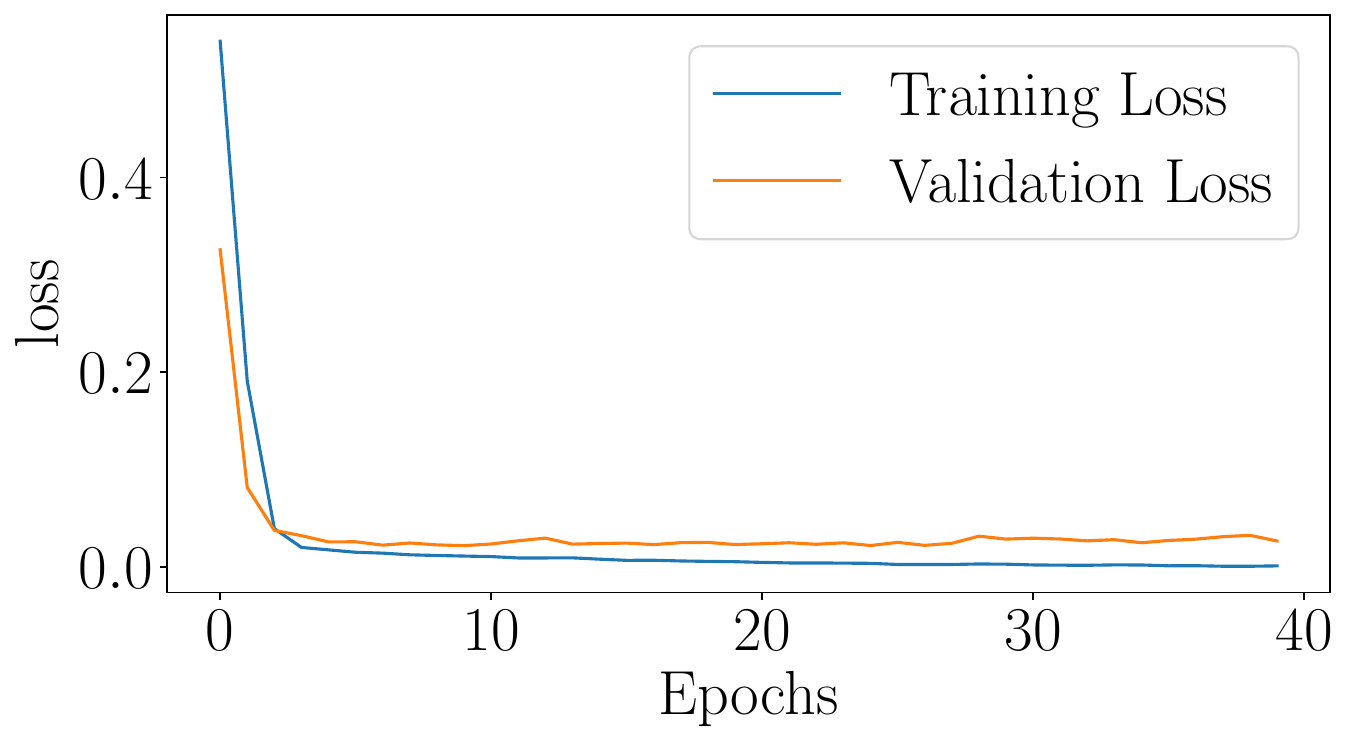}
        }
    \hfill
    \subfloat[]{%
        \includegraphics[width=0.38\textwidth]{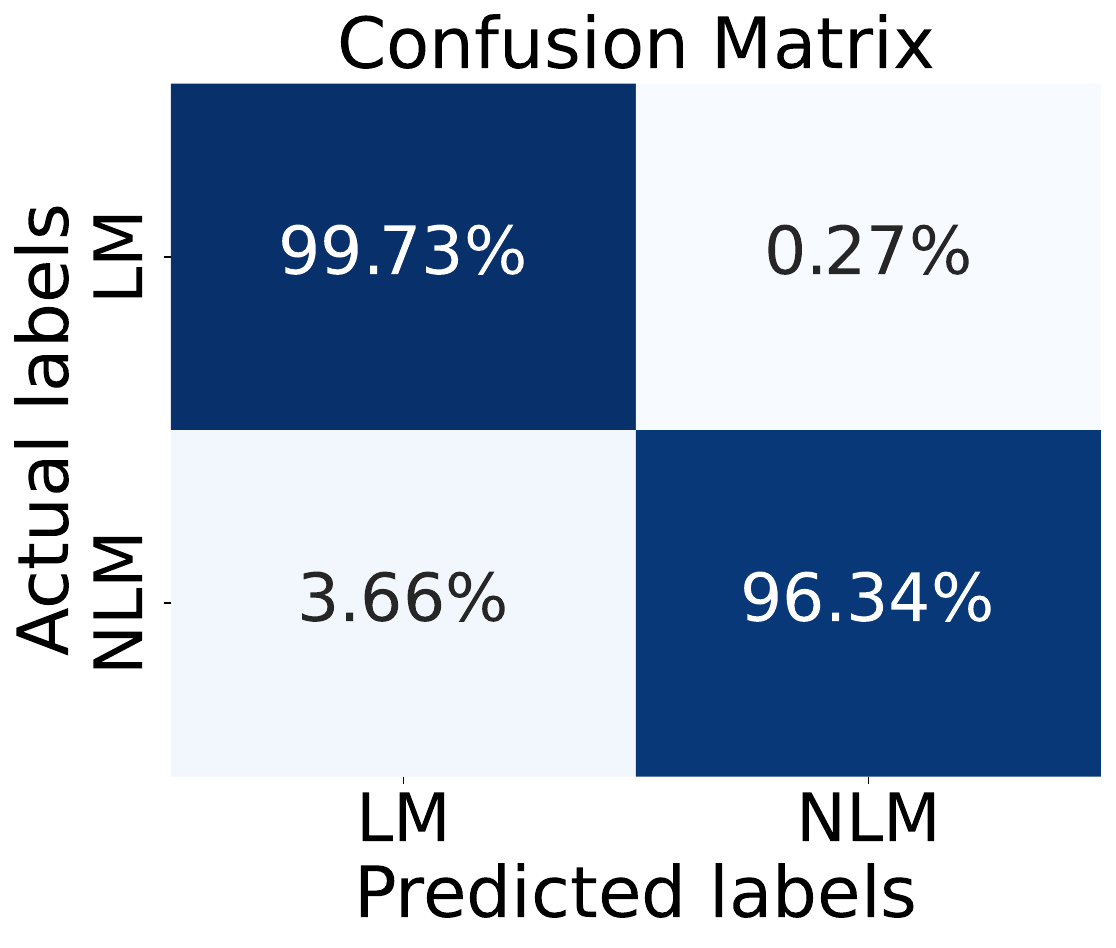}
        }
\caption{(a) Training (blue) and validation (orange) loss versus the number of epochs in case 1. (b) Average confusion matrix for case 1.}
\label{fig:model_train}
\end{figure}

In the process of training our CNN model, both training and validation loss were computed to monitor the model's learning progress and its ability to generalize. The training proceed for a maximum of 200 epochs; however, an early stopping callback is implemented to avoid overfitting. The training time is notably brief, amounting to only 30.19 seconds. However, testing the model requires a duration of 0.6 seconds (Table~\ref{tab:table1}(B)). Figure~\ref{fig:model_train}(a) presents the training and validation losses over the training epochs.
The performance of the model is assessed using the evaluation metrics in Section~\ref{sec:evaluation} and is shown in Table~\ref{tab:metrics}. The model's performance evaluation produces an average accuracy of 0.99 and an F1-score value of 0.94 (Table \ref{tab:metrics}).
We present the corresponding average confusion matrix in Figure~\ref{fig:model_train}(b). Within the confusion matrix, the model predicts correctly 99.73\% of the LM instances and 96.34\% of the NLM ones. These percentage-based values illustrate the model's high precision and recall scores, and show its robustness in accurately classifying nodes within local and nonlocal regions for the training dataset.

\begin{figure}[tbh!] 
\centering
\subfloat[]{%
\includegraphics[width=0.4\textwidth]{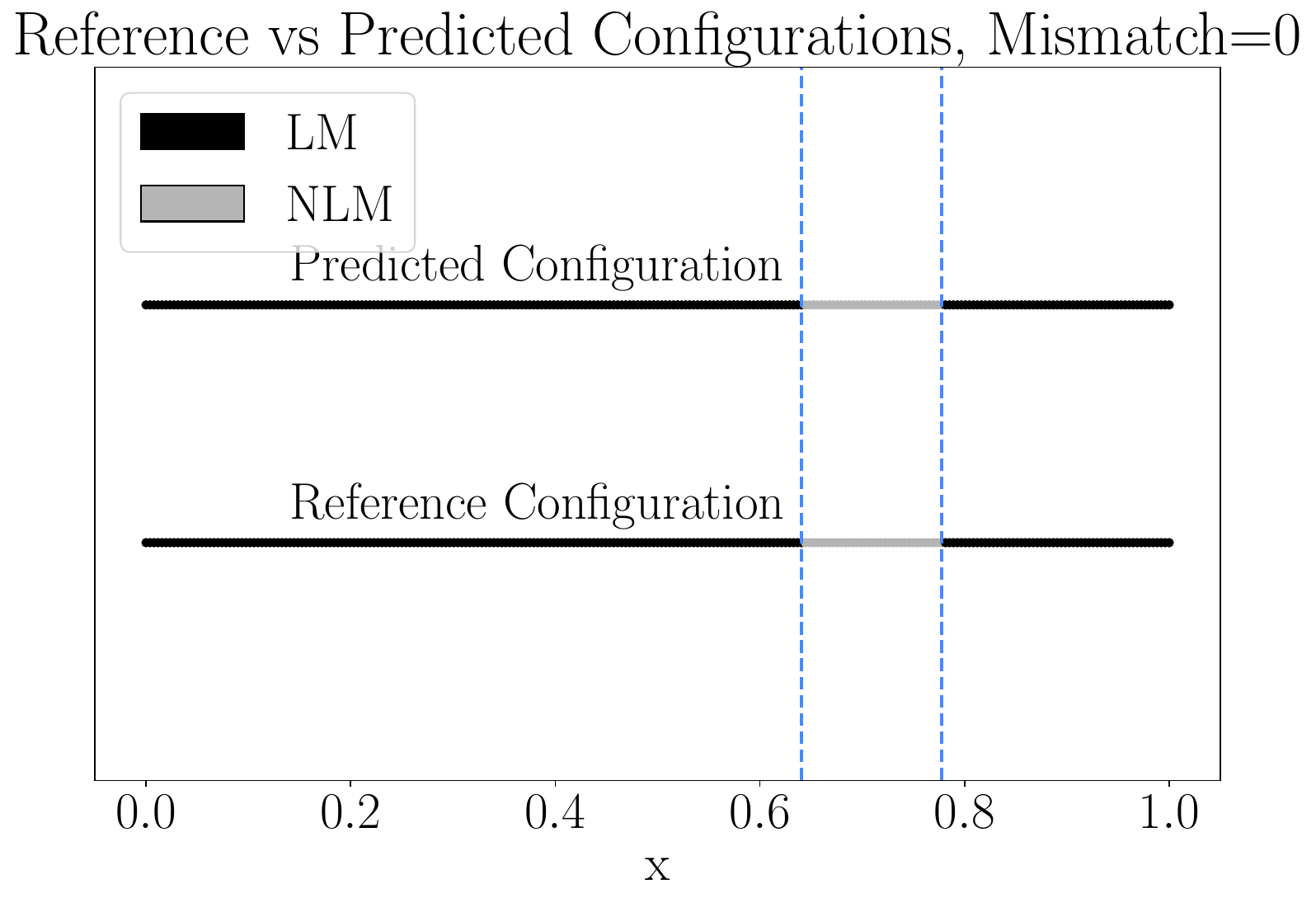}%
}%

\subfloat[]{%
\includegraphics[width=0.4\textwidth]{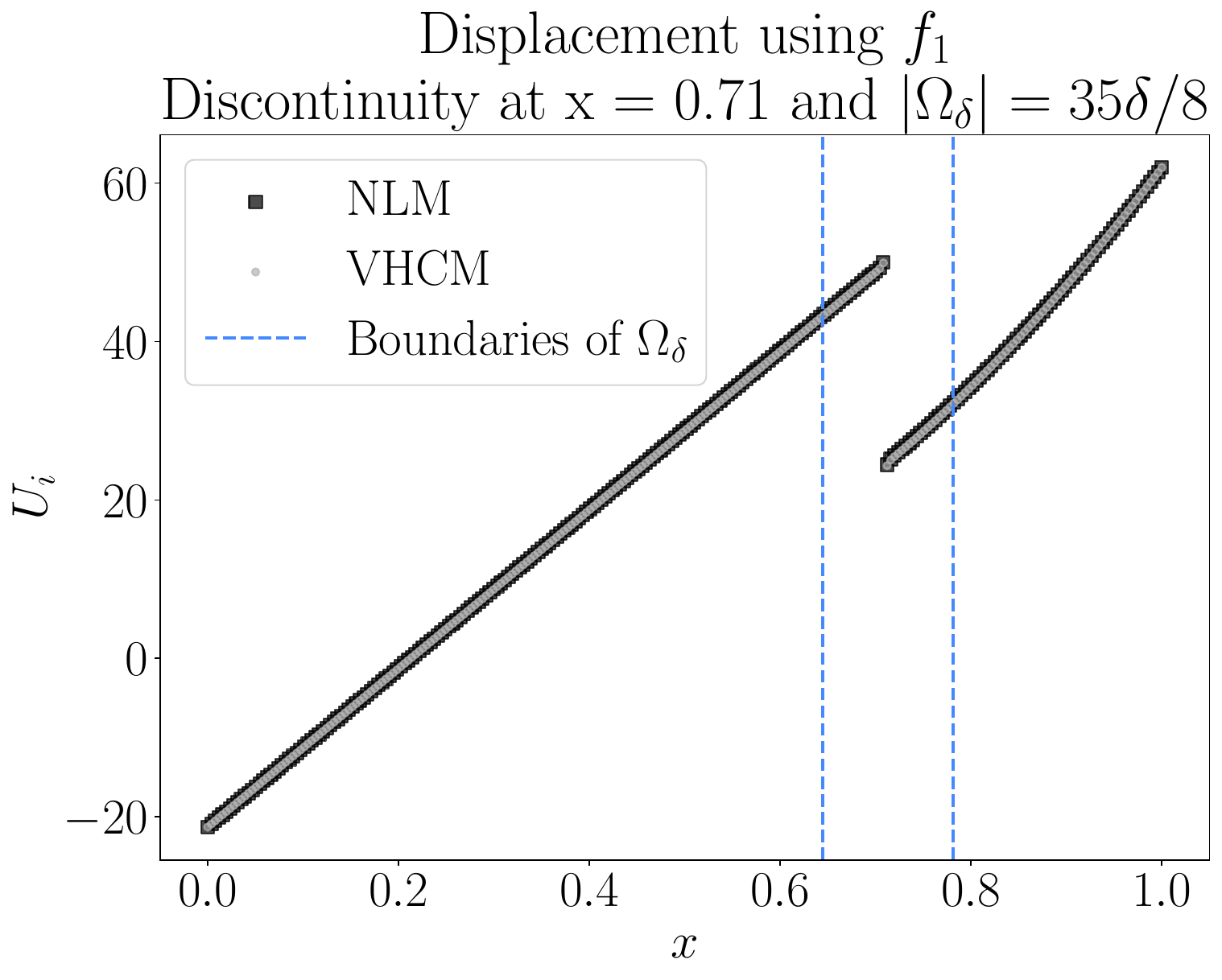}%
}%
\subfloat[]{%
\includegraphics[width=0.4\textwidth]{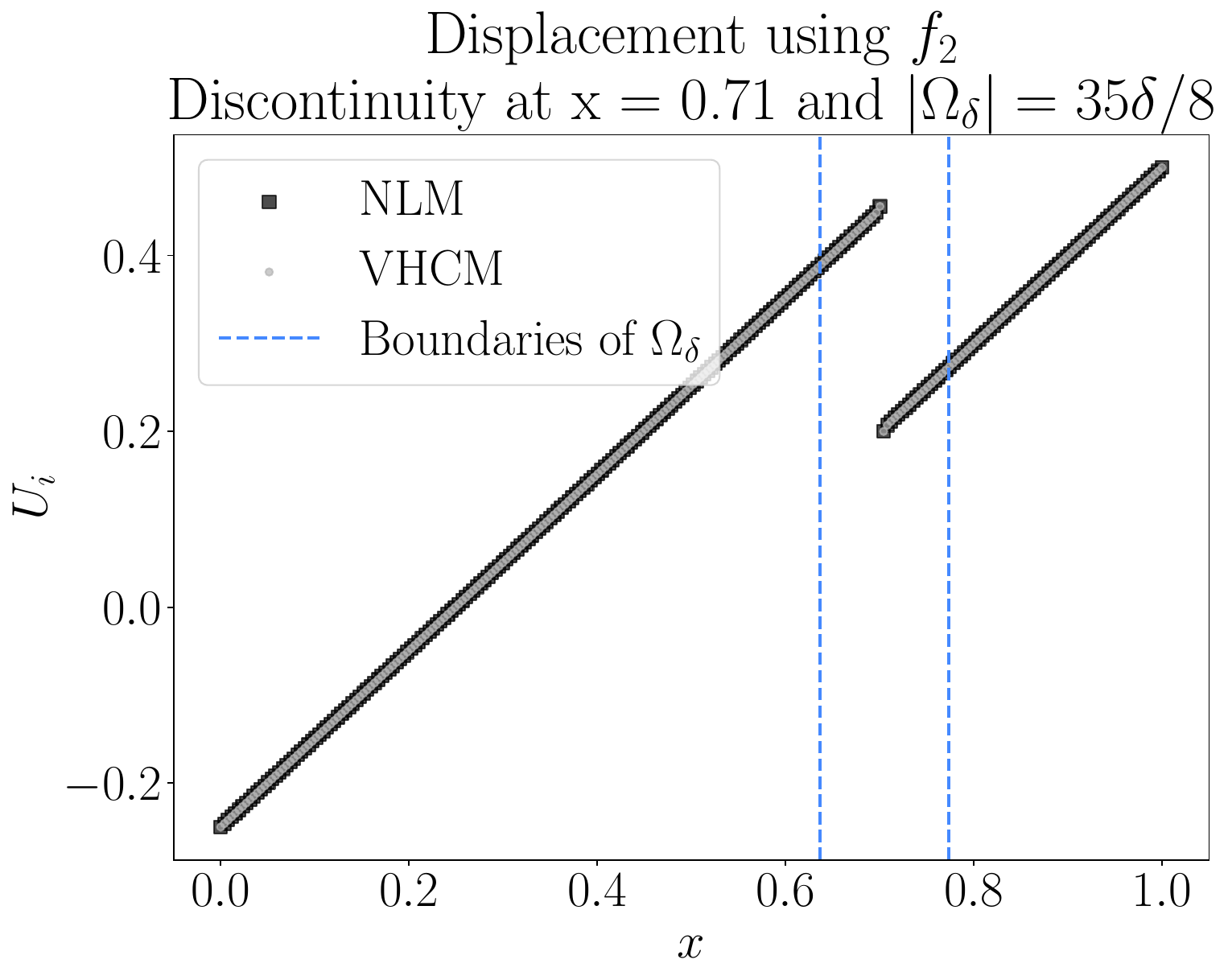}%
}%
\caption{Error estimation after prediction with discontinuity at $x=0.71$. (a) Comparison between the reference and predicted configurations of local and nonlocal regions. (b-c) Peridynamic region configuration after prediction with discontinuity at $x=0.71$.  $u_{\text{NLM}}$ is represented by \textcolor{black}{$\blacksquare$} and $u_{\text{VHCM}}$ is represented by~\textcolor{gray}{$\bullet$}, (b) displacement fields $u_\text{NLM}$ and $u_\text{VHCM}$ using load $f_1$. $\mathcal E(u_\text{NLM},u_\text{VHCM})=5.5282 \times 10^{-5}$; (c) displacement fields $u_\text{NLM}$ and $u_\text{VHCM}$ using load $f_2$. $\mathcal E(u_\text{NLM},u_\text{VHCM})=5.5285 \times 10^{-5}$.}
\label{fig:after_prediction}
\end{figure}

To evaluate the results after regions detection, we estimate the error in Eq~\eqref{eq:l2_norm} between the fully nonlocal displacement $u_\text{NLM}$ and the coupled displacement $u_\text{VHCM}$, whose nonlocal region is the one predicted by the CNN. An example of the results for both $f_1$ and $f_2$ is reported in  Figure~\ref{fig:after_prediction}, where, in both cases, the discontinuity is at $x=0.71$. The value of $\mathcal E(u_\text{NLM},u_\text{VHCM})$ using $f_1$ is $5.5282 \times 10^{-5}$; while using $f_2$ is $5.5285 \times 10^{-5}$.

For the test set, the predicted nonlocal region consistently induces an error in the coupled solution much lower than the tolerance used during training . Across all cases, the estimated errors fall within the range of \( 2.18 \times 10^{-6} \) to \( 5.32 \times 10^{-3} \) using $f_1$ and \( 2.17 \times 10^{-6} \) to \( 1.36 \times 10^{-4} \) using $f_2$ (Table \ref{tab:metrics}).

\begin{table}[tbp!]
\centering
\caption{CNN evaluation metrics and error summary for the test set.}
\label{tab:metrics}
\begin{tabular}{lcccc}
\toprule
 & \textbf{Accuracy} & \textbf{F1-score} & \textbf{Load} & \textbf{$\mathcal{E}(u_{\text{NLM}}, u_{\text{VHCM}})$} \\
\midrule
\makecell{\textbf{Case 1}:\\  Full-domain Data} 
 & 0.99 & 0.94 & \(f_1\) & \(2.18 \times 10^{-6}\) -- \(5.32 \times 10^{-3}\) \\
& & &  \(f_2\) & \(2.17 \times 10^{-6}\) -- \(1.36 \times 10^{-4}\) \\
  & & & \(f_4\)(\(t=0.05\))&\({8.8 \times 10^{-4}}^{\text{*}}\) \\
  & & & \(f_4\)(\(t=0.0005\))&\({1.03 \times 10^{-3}}^{\text{*}}\) \\
  & & &\(f_5\) & \({8.2 \times 10^{-4}
}^{\text{*}}\) \\
\addlinespace
\midrule
\addlinespace
\makecell{\textbf{Case 2}:\\ Windowing data}
 & 0.96 & 0.97 & \(f_1\) & \(3.50 \times 10^{-6}\) -- \(2.34 \times 10^{-3}\)
 \\
  & & & \(f_2\)& \(3.49 \times 10^{-6}\) -- \(1.97 \times 10^{-5}\) \\
  & & & \(f_4\) (\(t=0.05\))& \(5.30 \times 10^{-5}\) \\
 & & & \(f_4\) (\(t=0.0005\)) & \(2.55 \times 10^{-5}\) \\
  & & & \(f_5\)& \(1.30 \times 10^{-4}\) \\
\bottomrule
\end{tabular}

\smallskip
\raggedright
\footnotesize \(^*\)Here the coupled solution $u_\text{VHCM}$ corresponds to the fully local solution $u_\text{LM}$. The estimated error is then computed as the difference between $u_\text{NLM}$ and $u_\text{LM}$.
\end{table}

\subsection{Case 2: Window-based input data}
\label{sec:numerical:results:classification}

The results in the previous Section~\ref{sec:numerical:results:fulldomain} showed promise that machine learning could support the selection of local-nonlocal coupling regions; however, the full-load strategy does not have satisfactory generalization properties when testing using unseen data. In this section, we illustrate how the windowing approach circumvents these limitations.

\begin{figure}[tb!]
\centering
     \subfloat[]{%
        \includegraphics[width=0.6\textwidth]{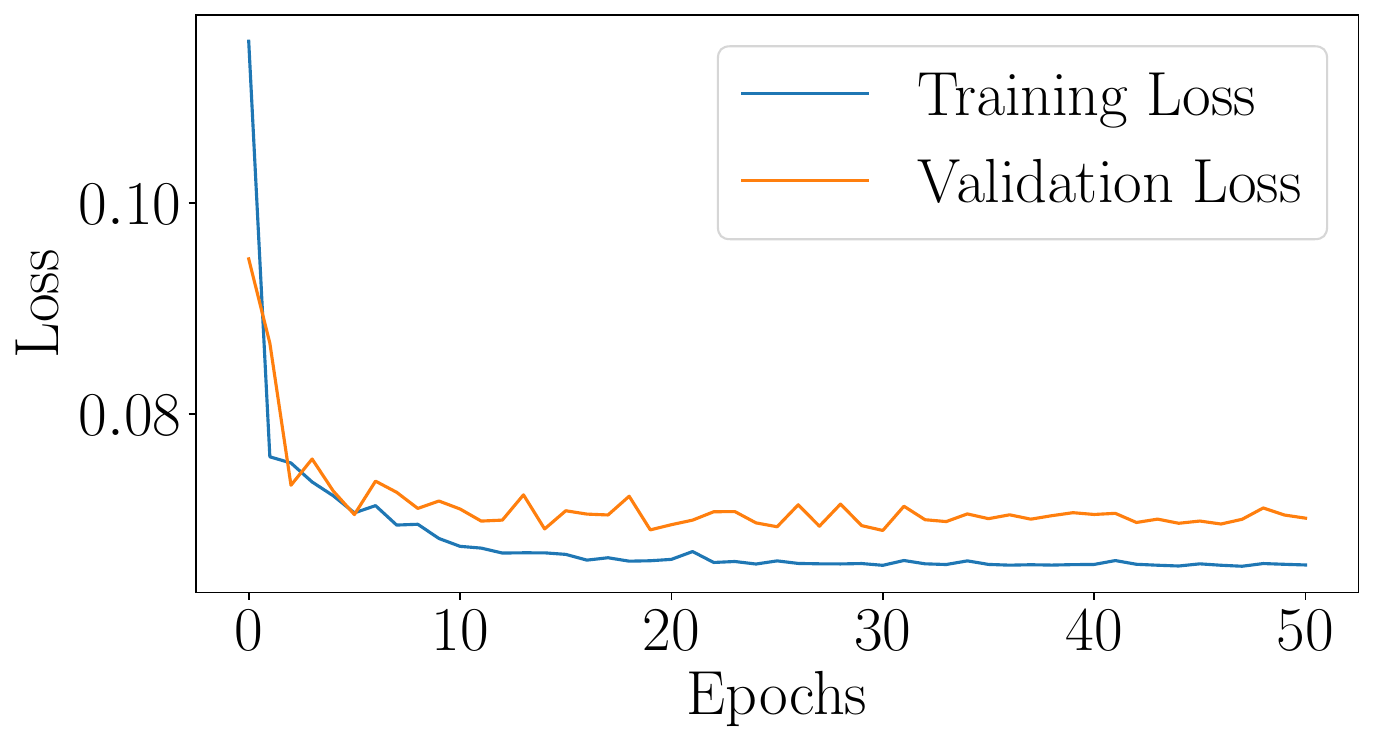}%
        }%
         \subfloat[]{%
        \includegraphics[width=0.4\textwidth]{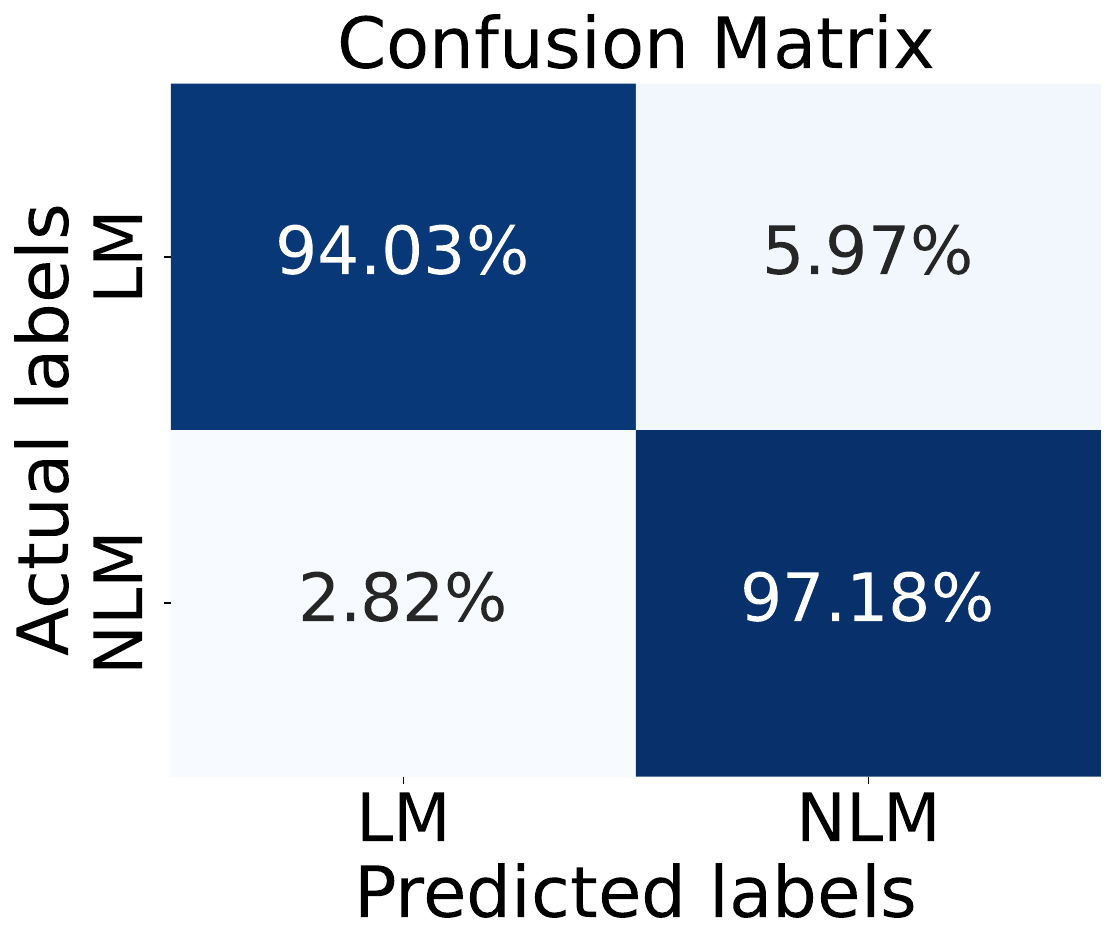}%
        }%
\caption{(a) Training (blue) and validation (orange) loss versus the number of epochs in case 2. (b) Confusion matrix for windowing dataset in case 2.}
\label{fig:confusion_matrix}
\end{figure}

As before, the number of epochs is set to of 200; however, an early stopping callback is used to avoid overfitting.  Training the model with windowed data take 137.17 seconds while testing requires a duration of 10.9 seconds (Table~\ref{tab:table1}(B)). Figure \ref{fig:confusion_matrix}(a) presents the training and validation losses over the training epochs. While both losses decrease overall, the close tracking between training and validation loss suggests that the model is not overfitting to the training data. The plot shows that the callback effectively prevented overfitting by stopping training at epoch 50 when the validation loss showed no further improvement.
The performance of the model was assessed using the evaluation metrics are shown in Table~\ref{tab:metrics}. The performance metrics show an accuracy of 0.96 and an F1-score of 0.97;
the corresponding confusion matrix, reported in Figure~\ref{fig:confusion_matrix}(c),  shows that the model predicted correctly 94.03\% of the LM instances and 97.18\% of of the NLM ones. Also in this case, the model has high precision and recall scores.

For the test set, the predicted nonlocal region induces an error in the coupled solution much lower than the tolerance $\epsilon$ used during training all cases using both $f_1$ and $f_2$. Across all cases, the estimated errors fall within the range of \( 3.50 \times 10^{-6} \) to \( 2.34 \times 10^{-3}\) using $f_1$ and \( 3.49 \times 10^{-6} \) to \( 1.97 \times 10^{-5}\) using $f_2$ (Table \ref{tab:metrics}). The lowest errors are obtained when using the original version of loads $f_1$ and $f_2$, which means without transformations. 

\begin{figure}[tbp!]
 \centering
     \subfloat[]{%
        \includegraphics[width=0.5\textwidth]{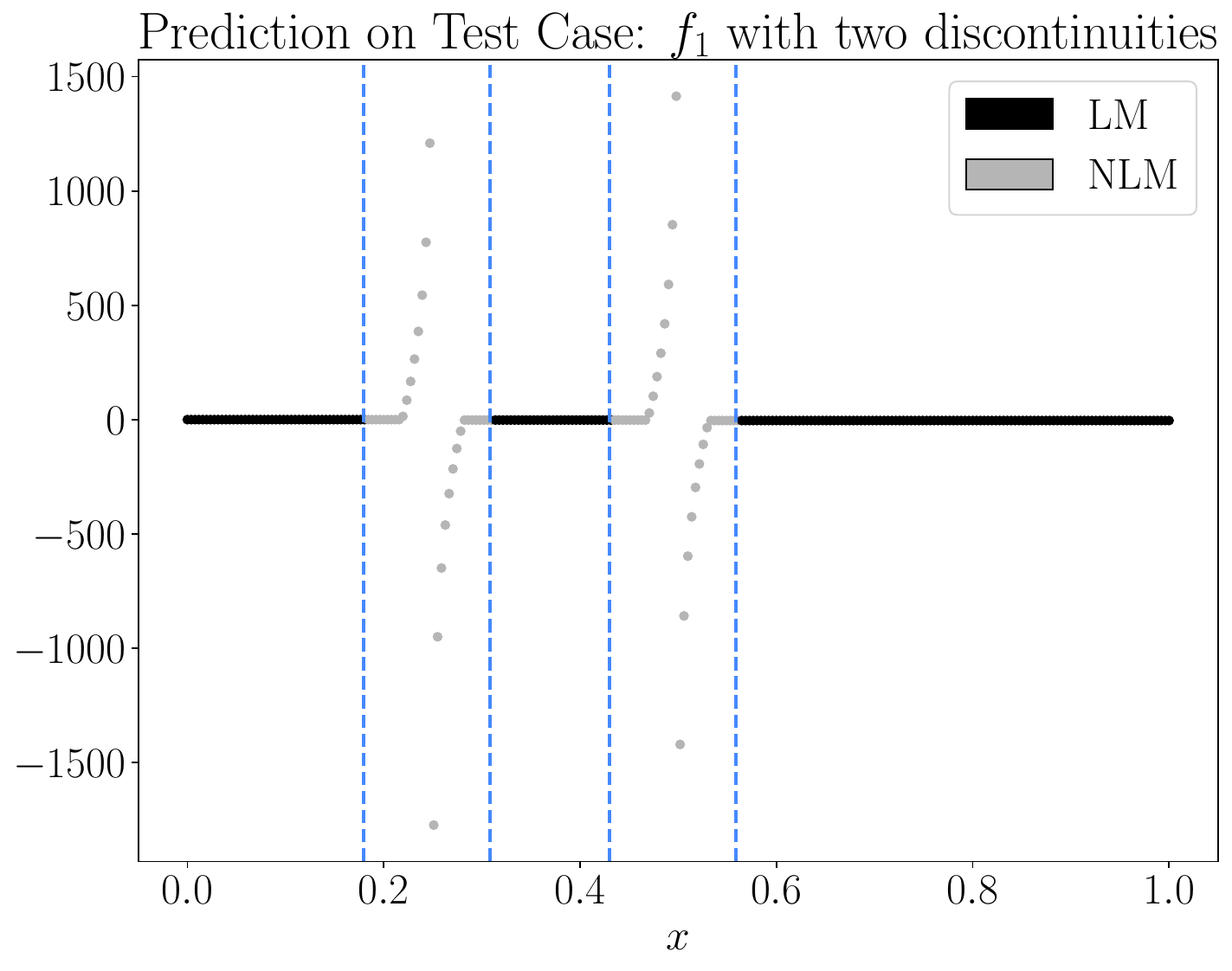}%
        }%
         \subfloat[]{%
        \includegraphics[width=0.5\textwidth]{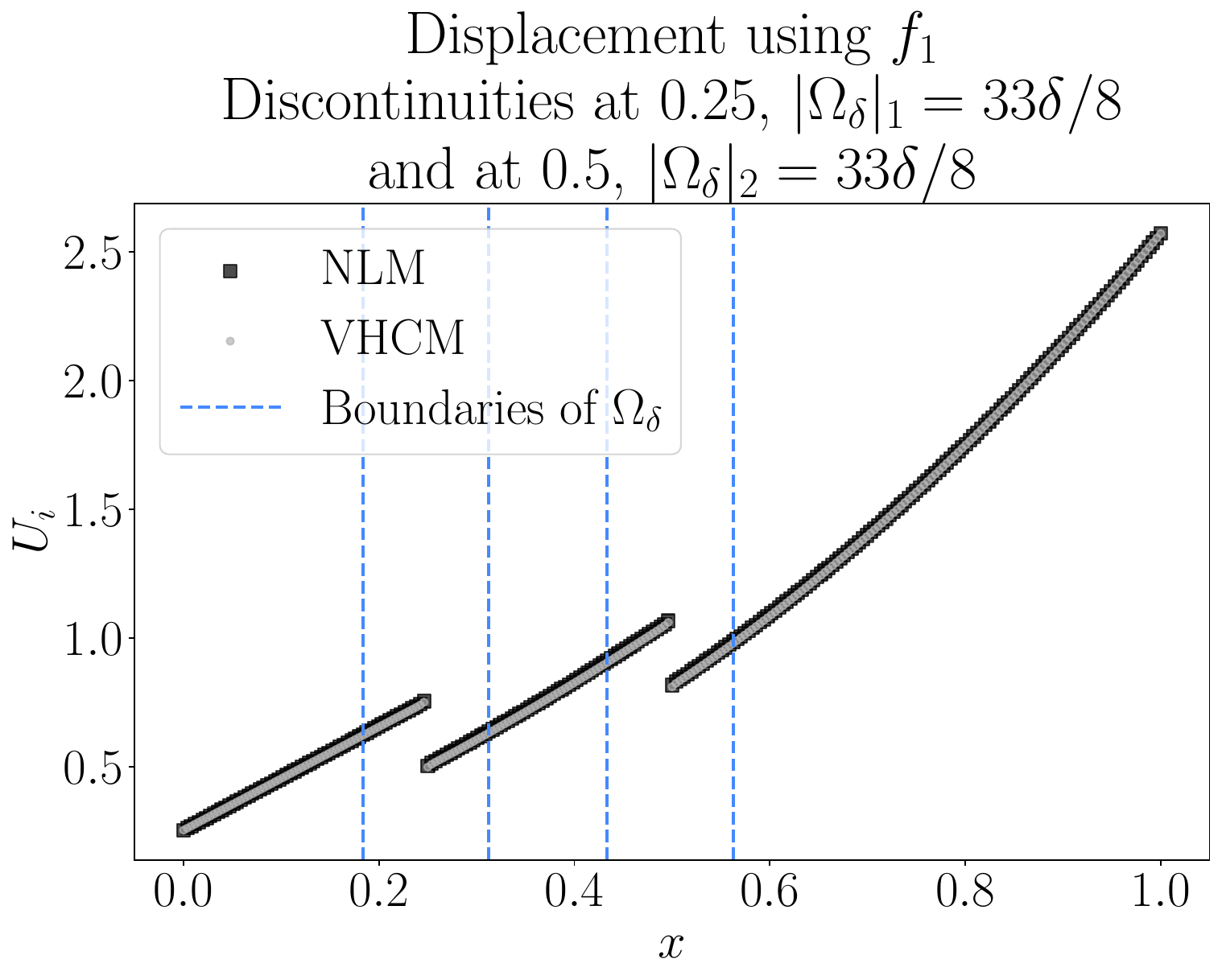}%
        }%
         \caption{Nonlocal region detection and error estimation after prediction for test case with two jumps using $f_1$. (a) Predicted configuration of local and nonlocal regions. The nodes located in the region where the local model will be used are labeled LM and represented in black. The nodes located in the region where the nonlocal model will be employed are labeled NLM and represented in light grey. (b) Displacement fields $u_{\text{NLM}}$, represented by \textcolor{black}{$\blacksquare$} and $u_{\text{VHCM}}$, represented by~\textcolor{gray}{$\bullet$}. The corresponding error is $3.299 \times 10^{-3}$.}
    \label{fig:test_two_jumps}
\end{figure}
To evaluate the windowing approach's ability to identify multiple discontinuities, we test our pre-trained model, which has not been trained on loads with more than one discontinuity, using a load with two discontinuities. The results are presented in Figure \ref{fig:test_two_jumps}. The predicted configurations of local and nonlocal regions are shown in Figure~\ref{fig:test_two_jumps}(a), while the displacements $u_{\text{NLM}}$ (represented by \textcolor{black}{$\blacksquare$}) and $u_{\text{VHCM}}$ (represented by~\textcolor{gray}{$\bullet$}) are shown in Figure~\ref{fig:test_two_jumps}(b). The corresponding error, denoted as $\mathcal E(u_{\text{NLM}}, u_\text{VHCM})$, is $3.299 \times 10^{-3}$.


In addition, we test our model using load $f_4$ (Eq. \ref{eq:force_f4}) and $f_5$ (Eq. \ref{eq:force_f5}); these loading functions do not belong to the training set. Thus, with these examples, we test mild generalization properties of the algorithm. 

\begin{figure}[ptb!] 
\centering
    \subfloat[t=0.05]{%
        \includegraphics[width=0.4\textwidth]{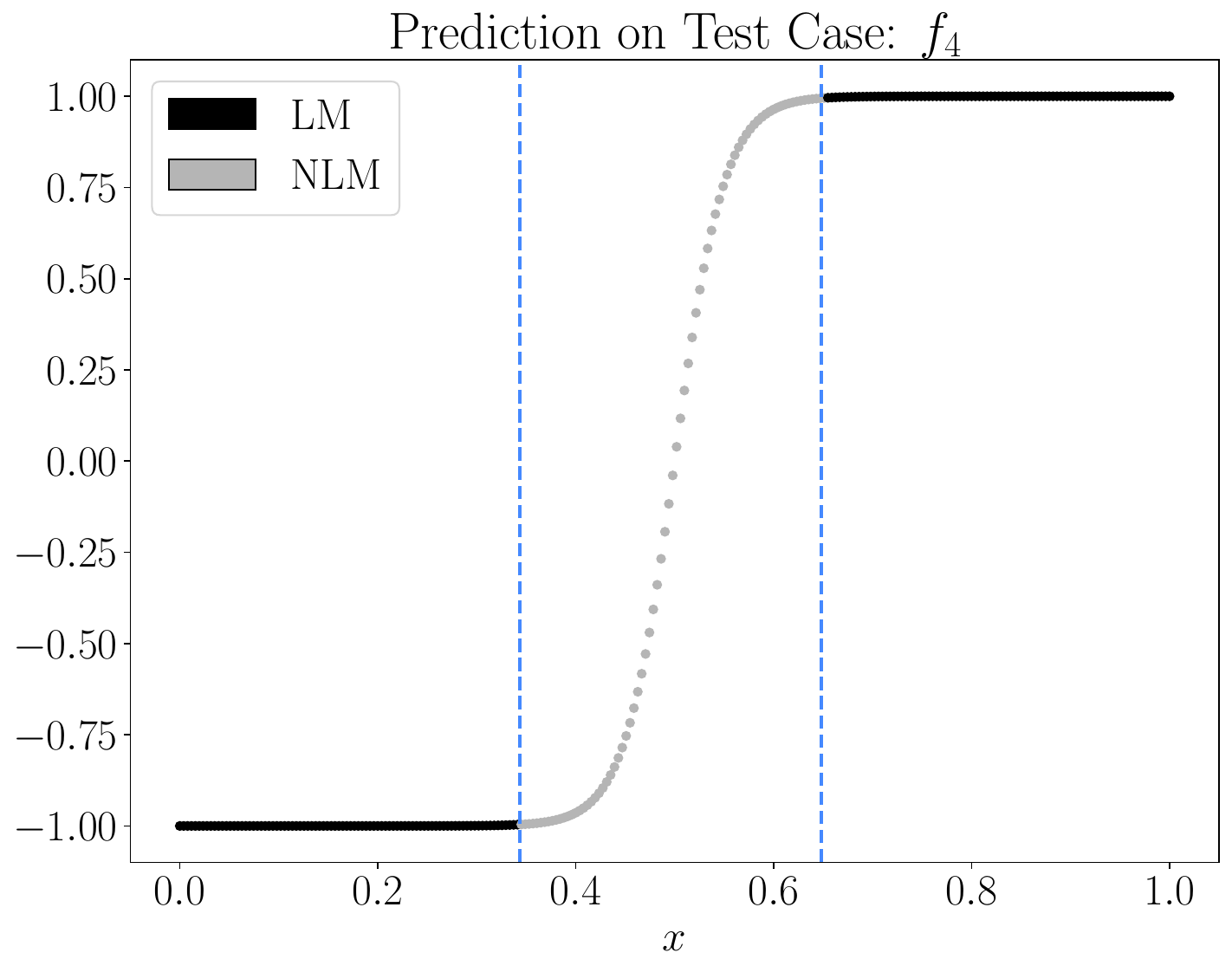}%
        }%
\hspace{0.05\textwidth}
    \subfloat[t=0.05]{%
        \includegraphics[width=0.4\textwidth]{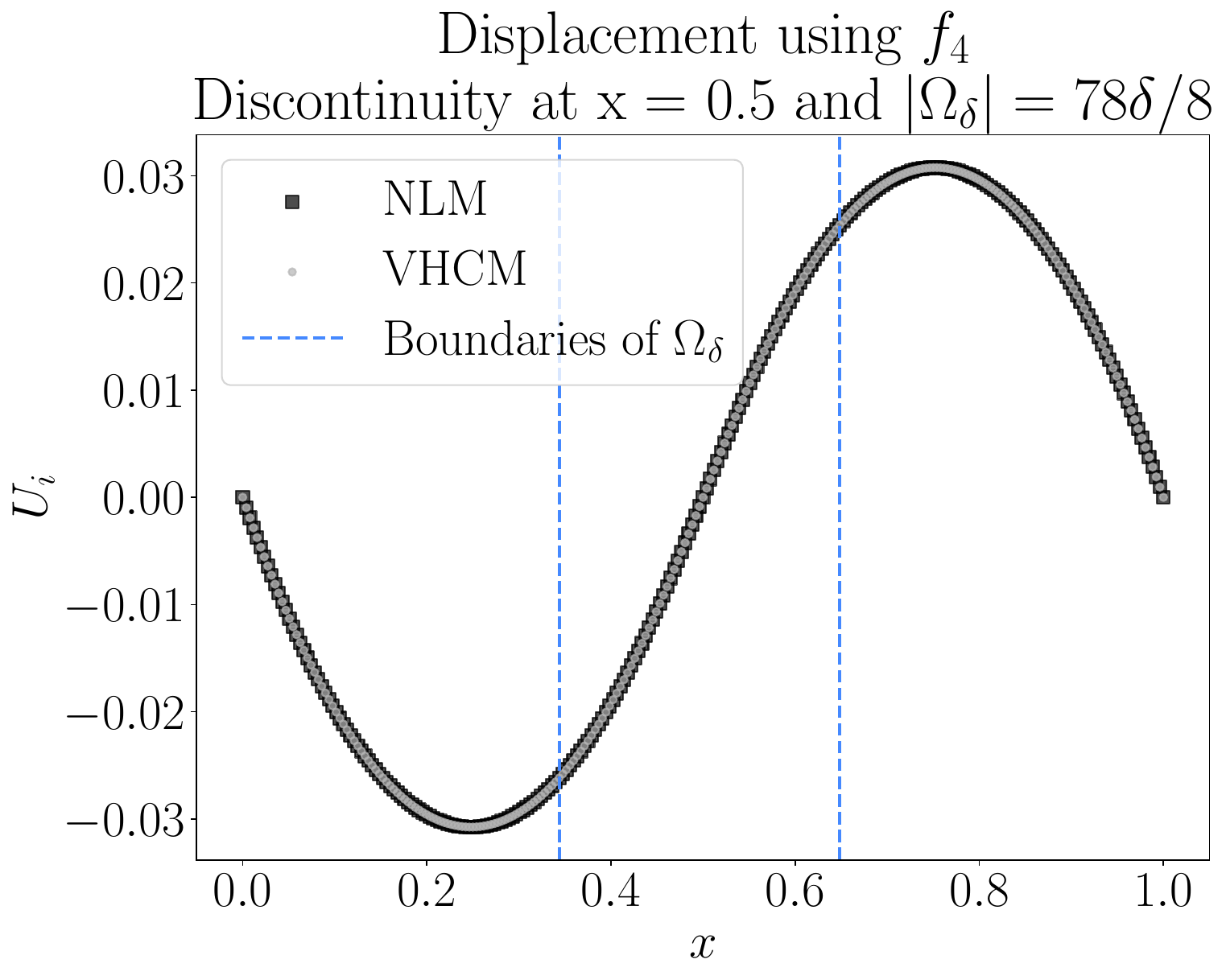}%
        }
\vspace{0.01cm}
 \subfloat[t=0.0005]{%
        \includegraphics[width=0.4\textwidth]{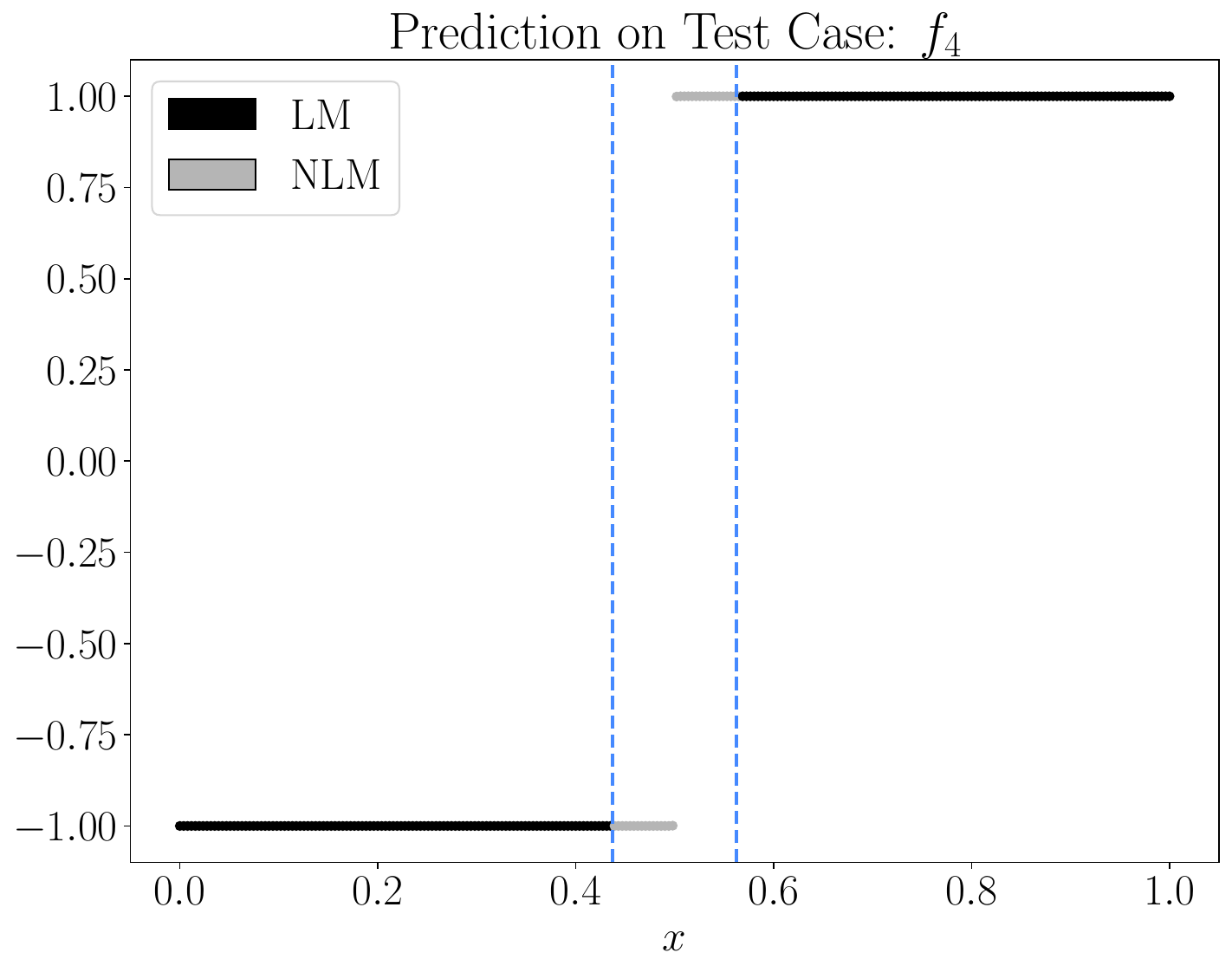}%
        \label{fig:c}%
        }%
\hspace{0.05\textwidth}
    \subfloat[t=0.0005]{%
        \includegraphics[width=0.4\textwidth]{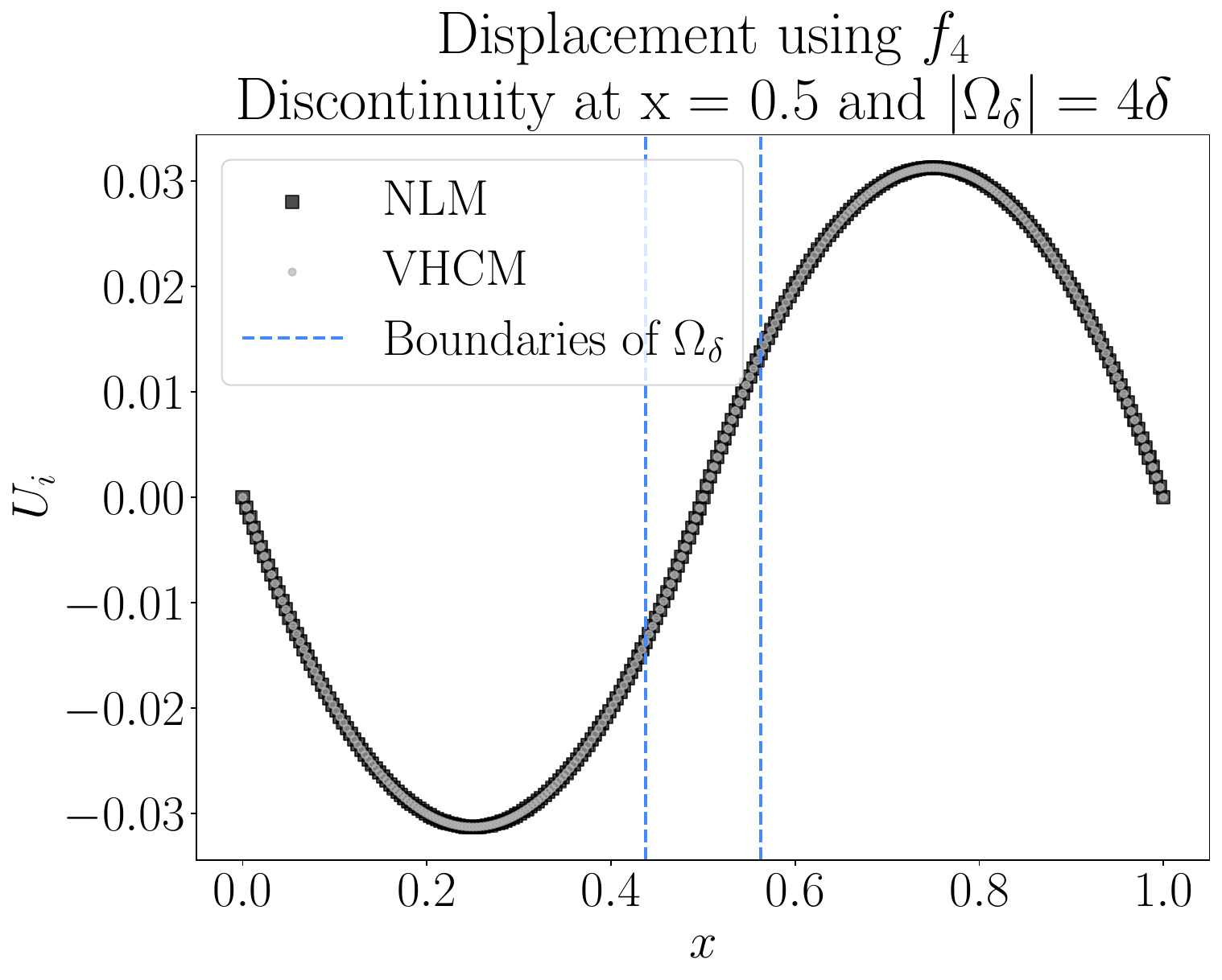}%
        \label{fig:d}%
        }%
\vspace{0.01cm}
        \subfloat[]{%
        \includegraphics[width=0.4\textwidth]{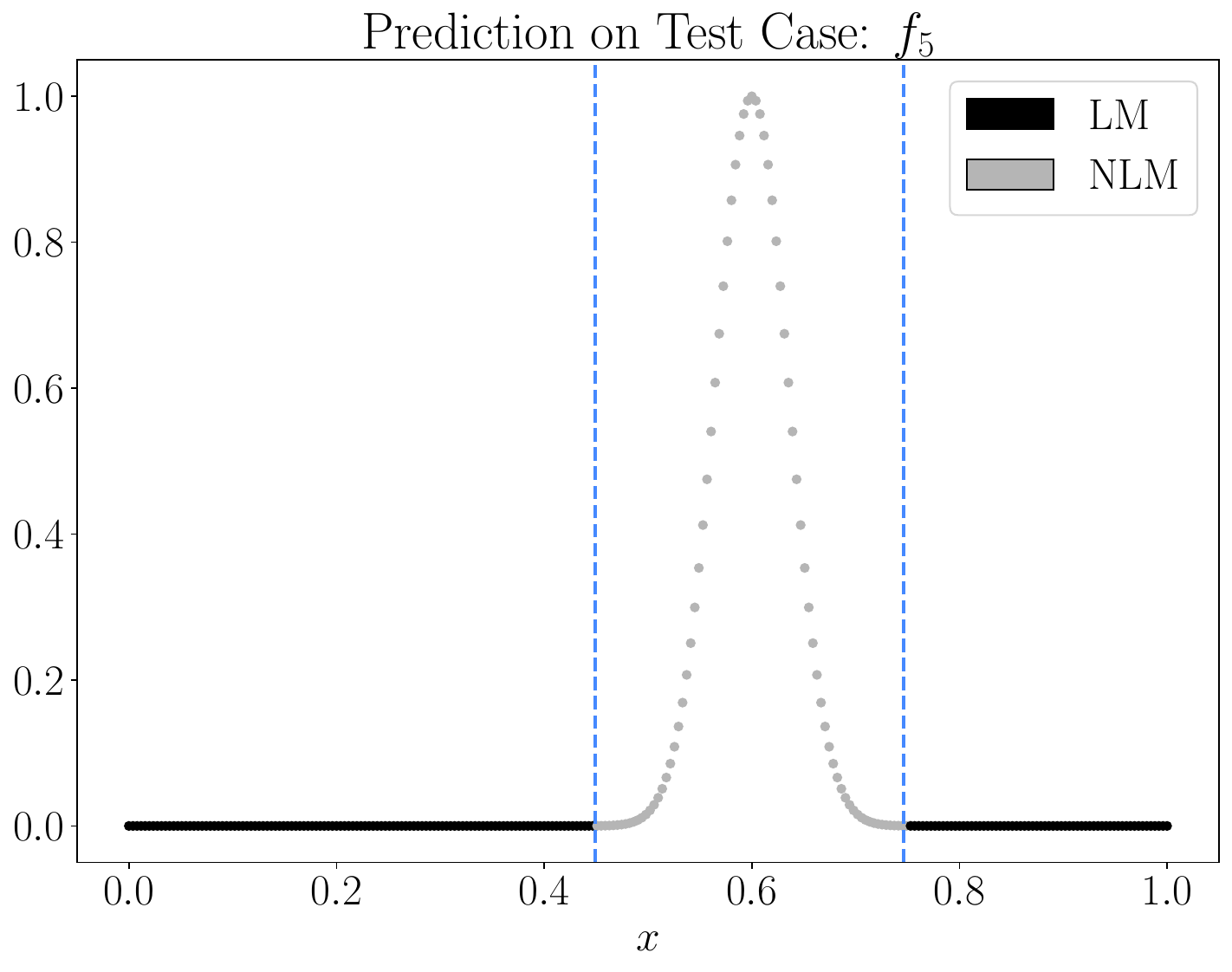}%
        \label{fig:e}%
        }%
\hspace{0.05\textwidth}
    \subfloat[]{%
        \includegraphics[width=0.4\textwidth]{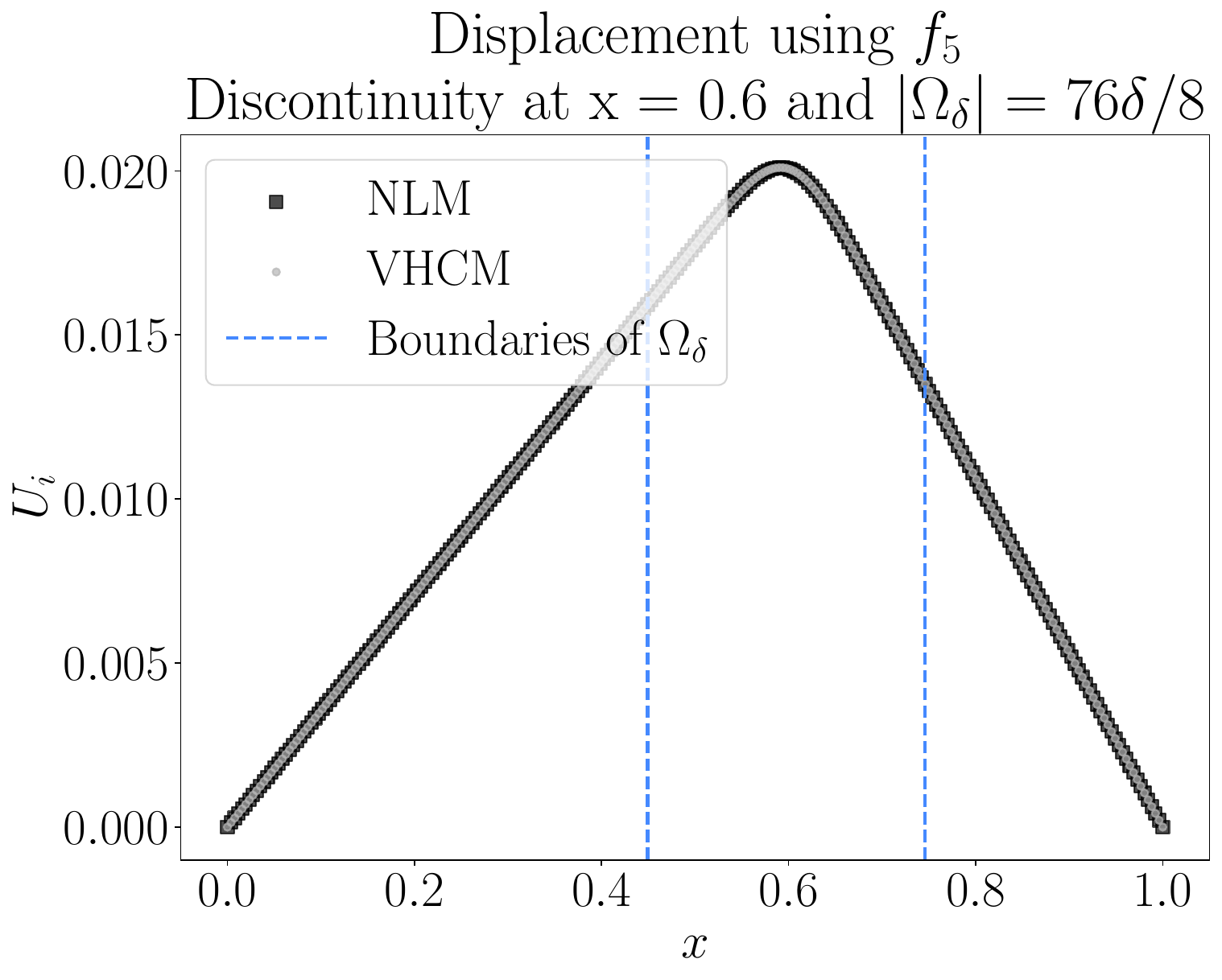}%
        \label{fig:f}%

        }%
    \caption{Nonlocal region detection and error estimation after prediction for general test cases $f_4$ and $f_5$ in Eq.~\eqref{eq:force_f4} and Eq.~\eqref{eq:force_f5}. (Left) Predicted configuration of local and nonlocal regions. The nodes located in the region where the local model will be used are labeled LM and represented in black. The nodes located in the region where the nonlocal model will be employed are labeled NLM and represented in light grey. (Right) Displacement fields $u_{\text{NLM}}$, represented by \textcolor{black}{$\blacksquare$} and $u_{\text{VHCM}}$, represented by~\textcolor{gray}{$\bullet$}. (a-b) Results for $f_4$ with $t=0.05$, the corresponding error is $5.30 \times 10^{-5}$. (c-d) Results for $f_4$ with $t=0.0005$, the corresponding error is $2.55 \times 10^{-5}$. (e-f) Results for $f_5$, the corresponding error is $1.30 \times 10^{-4}$.}
\label{fig:test_case_error}
\end{figure}

For $f_4$, we consider two scenarios; one with t=0.05 and another with t=0.0005. In all cases, the second derivative is estimated and subsequently segmented into windows, serving as the input for the pre-trained model. Once again, to evaluate the precision of the region detection, we estimate the error between the fully nonlocal displacement ($u_{\text{NLM}}$) and the coupled displacement ($u_\text{VHCM}$) using Eq.~\eqref{eq:l2_norm}. Figure~\ref{fig:test_case_error}(Left) shows the predicted configurations of local and nonlocal regions. Figure~\ref{fig:test_case_error}(Right) shows $u_{\text{NLM}}$, represented by \textcolor{black}{$\blacksquare$} and $u_{\text{VHCM}}$, represented by~\textcolor{gray}{$\bullet$}.

For $f_4$ with t=0.05, the prediction labels and configurations of local and nonlocal regions are showcased in Figure~\ref{fig:test_case_error}(a-b). The corresponding error value, denoted as $\mathcal E(u_{\text{NLM}}, u_\text{VHCM})$, is $5.30 \times 10^{-5}$. When testing $f_4$ with t=0.0005, the findings are similarly reported in Figures~\ref{fig:test_case_error}(c-d), with an error value of $2.55 \times 10^{-5}$.

For $f_5$, as defined in Eq. \ref{eq:force_f5}, we follow the same process. The results are presented in Figure~\ref{fig:test_case_error}(e-f), and the corresponding error $\mathcal E(u_{\text{NLM}}, u_\text{VHCM})$ is $1.30 \times 10^{-4}$.

\section{Conclusion}
\label{sec:conclusion}
This paper shows a proof of concept for a ML-based identification of local-nonlocal coupling regions that is independent of the local and nonlocal models utilized. Specifically, the load used as identification input is independent of the two models and the method used to generate local and nonlocal solutions is irrelevant. Furthermore, while we only considered a non-overlapping approach, a coupling approach with overlapping regions could also be used (examples can be found in ~\cite{diehl2022coupling}). 

Among the two approaches we presented, the full-domain approach yields promising results for the classification of the local and nonlocal regions as long as the solutions do not feature multiple discontinuities. To overcome this limitation, we introduced the windowing approach based on a node-wise classification. The latter shows promising results even in the presence of several discontinuities without retraining (see Section~\ref{sec:numerical:results:classification}). This model demonstrated robust performance, as reflected by high evaluation metrics, with an accuracy of 0.98 and an F1-score of 0.97. The results were further validated by the corresponding confusion matrix (see Figure~\ref{fig:confusion_matrix}(c)), which provides a detailed breakdown of the model's classification accuracy.
Nevertheless, a minor limitation in our model's predictive precision is noticed. Specifically, our analysis identified a misclassification rate of 5.21\% of instances as LM (nodes in Local region) and 1.14\% as NLM (nodes in nonlocal region). These relatively low values of the total predictions highlight areas of potential improvement.


The proof of concept approach introduced in this work opens several avenues of improvement and extension. In the results presented in this study, the horizon $\delta=m\cdot h$ is constant; for more realistic tests, varying values should be investigated (the work in~\cite{diehl2016bond} can be used as a guide for the choice of $h$ and $\delta$). As for the boundary conditions, we use homogeneous Dirichlet boundary conditions, while in real-world applications mixed conditions are usually prescribed. More importantly, an extension to two- and three-dimensional problems is mandatory and it is the subject of our current work. Furthermore, stress testing the approach by using more complex forcing functions from a larger subspace should also be explored (e.g. randomly generated functions with prescribed regularity properties).

Another relevant aspect not directly addressed in this work is the presence of cracks and fractures. These features were mimicked in the current study by the presence of discontinuities. In two- and three-dimensional settings, we plan to incorporate these features into the machine learning model and validate our results against real experiments.

In terms of machine learning algorithms, the windowing approach showed promising results and is the candidate strategy to proceed with in higher dimensions. However, the open question is the choice of the shape of the window. Possible candidates in two dimensions could be rectangles or ellipses. More generally, the shape of the window should be chosen such that it matches the support of the nonlocal kernel function. Alternatively, one could parameterize a family of two-dimensional shapes, in which case the parameters would become learnable parameters in the training process. Another open question is the architecture of the CNN for higher dimensions as the input layer and output layer need to be adapted to the higher dimensions.

Additionally, due to the fact that CNNs fail in the presence of nonuniform unstructured grids, our future exploration includes the use of graph neural networks (GNNs)~\cite{zhou2020graph,wu2020comprehensive, gladstone2023gnn}. Given the graph-like structure of our problem, leveraging GNNs could offer a compelling approach to model and predict complex interactions within nodes in local and nonlocal regions. The adaptability of GNNs to capture relationships among various regions' nodes makes them a promising approach especially for higher dimensions.

\section{Funding statement}

Sandia National Laboratories is a multimission laboratory managed and operated by National Technology \& Engineering Solutions of Sandia, LLC, a wholly owned subsidiary of Honeywell International Inc., for the U.S. Department of Energy’s National Nuclear Security Administration under contract DE-NA0003525.

This paper describes objective technical results and analysis. Any subjective views or opinions that might be expressed in the paper do not necessarily represent the views of the U.S. Department of Energy or the United States Government.

\bibliographystyle{abbrv}
\bibliography{template}  






\end{document}